%% file: main.tex
\pdfoutput=1
\documentclass{article}
\PassOptionsToPackage{dvipsnames,svgnames,x11names}{xcolor}
\usepackage{iclr2027_conference,times}

\usepackage[utf8]{inputenc}
\usepackage[T1]{fontenc}
\usepackage[american]{babel}
\usepackage{microtype}
\usepackage{amsmath}
\usepackage{amssymb}
\usepackage{amsfonts}
\usepackage{amsthm}
\usepackage{bbm}
\usepackage{bm}
\usepackage{nicefrac}
\usepackage{xcolor}
\usepackage{graphicx}
\usepackage{subcaption}
\usepackage{wrapfig}
\usepackage{booktabs}
\usepackage{multirow}
\usepackage{bigstrut}
\usepackage{colortbl}
\usepackage{diagbox}
\usepackage{tabularx}
\usepackage{algorithm}
\usepackage{algpseudocode}
\usepackage{enumitem}
\usepackage{pifont}
\usepackage{listings}
\usepackage{textcomp}
\usepackage{hyperref}
\usepackage{url}
\hypersetup{
  colorlinks=true,
  linkcolor=MidnightBlue,
  citecolor=MidnightBlue,
  urlcolor=MidnightBlue,
  pdftitle={MemAgent: Learning to Manage Heterogeneous Memory Providers for LLM Agents},
  pdfauthor={Yongxian Wei, Yilin Zhao, Runxi Cheng, Xinrui Chen, Chun Yuan, Yaoru Wang, Jiahong Yan, Dian Li}
}
\usepackage[capitalize,noabbrev]{cleveref}
\let\cite\citep

\crefname{section}{Sec.}{Secs.}
\Crefname{section}{Section}{Sections}
\Crefname{table}{Table}{Tables}
\crefname{equation}{Eq.}{Eqs.}
\crefname{algorithm}{Alg.}{Algs.}
\crefname{figure}{Fig.}{Figs.}
\crefname{appendix}{App.}{Apps.}
\graphicspath{{figure/}}

\newcommand{\romark}[1]{\uppercase\expandafter{\romannumeral #1\relax}}

\definecolor{OurBG}{rgb}{0.9,0.9,1.0}
\newcolumntype{C}{>{\centering\arraybackslash}X}
\newcommand{\ie}{{\emph{i.e.}}}
\newcommand{\eg}{{\emph{e.g.}}}

\makeatletter
\newcommand\figcaption{\def\@captype{figure}\caption}
\newcommand\tabcaption{\def\@captype{table}\caption}
\makeatother

\theoremstyle{plain}

\theoremstyle{definition}

\theoremstyle{remark}

\title{\parbox{\textwidth}{\centering
MemAgent: Learning to Manage\\
Heterogeneous Memory Providers\\
for LLM Agents}}
\author{
\makebox[\dimexpr\textwidth-2\tabcolsep-3pt\relax][c]{%
\begin{tabular}{@{}cccc@{}}
\textbf{Yongxian Wei}$^{1}$ & \textbf{Yilin Zhao}$^{2}$ & \textbf{Runxi Cheng}$^{1}$ & \textbf{Xinrui Chen}$^{1}$ \\
\textbf{Chun Yuan}$^{1}$ & \textbf{Yaoru Wang}$^{2}$ & \textbf{Jiahong Yan}$^{2}$ & \textbf{Dian Li}$^{2}$ \\[2pt]
\multicolumn{4}{c}{{\small\normalfont $^{1}$Tsinghua University \qquad $^{2}$Tencent}}
\end{tabular}%
}
}

\iclrfinalcopy

\begin{document}
\maketitle
\lhead{Preprint}

\begin{abstract}
Current agents remain largely stateless across tasks, limiting their ability to continually improve from prior interactions and making memory essential for long-horizon agentic behavior.
Existing memory methods seek to reuse past experience, but most rely on a single memory representation (\eg, trajectories, reflections, skills, structured knowledge) whose effectiveness varies across task distributions.
Rethinking this design space, we evaluate 13 memory methods and find that no single method generalizes across benchmarks, revealing the potential of \emph{managing} heterogeneous memory providers.
We formulate agent memory as a \emph{routing} problem in which a memory agent decides which memory provider to retrieve from, whether to inject short-term memory, and which providers should store the resulting experience.
Based on this perspective, we propose \textbf{MemAgent}, featuring a content-aware routing architecture and a training-data synthesis pipeline. The routing architecture combines content-aware probing before retrieval, short-term memory gating during execution, and selective multi-provider storage, while the training pipeline synthesizes phase-specific supervision for routing decisions.
Across GAIA, WebWalkerQA, and xBench-DS, MemAgent improves average accuracy by 10.0\% and outperforms every individual memory method across all three benchmarks. These gains come with less than 0.3\% routing overhead and a 12\% reduction in average task steps.
Our code is available \href{https://github.com/WalkerWorldPeace/MemAgent}{here}.
\end{abstract}

\input{section/intro.tex}

\input{section/related.tex}
\input{section/preliminary.tex}
\input{section/method.tex}
\input{section/experiment.tex}

\input{section/conclusion.tex}
\nocite{wei2025learning,wei2025openvocabulary,wei2025modeling,wei2026optmerge,wei2026closedform}

\begingroup
\small
\bibliography{ref}
\bibliographystyle{iclr2027_conference}
\endgroup

\clearpage
\appendix
\input{section/appendix.tex}

\end{document}

%% file: section/intro.tex
\section{Introduction}
\begin{figure*}[tbh]
    \centering
    \includegraphics[width=0.8\textwidth]{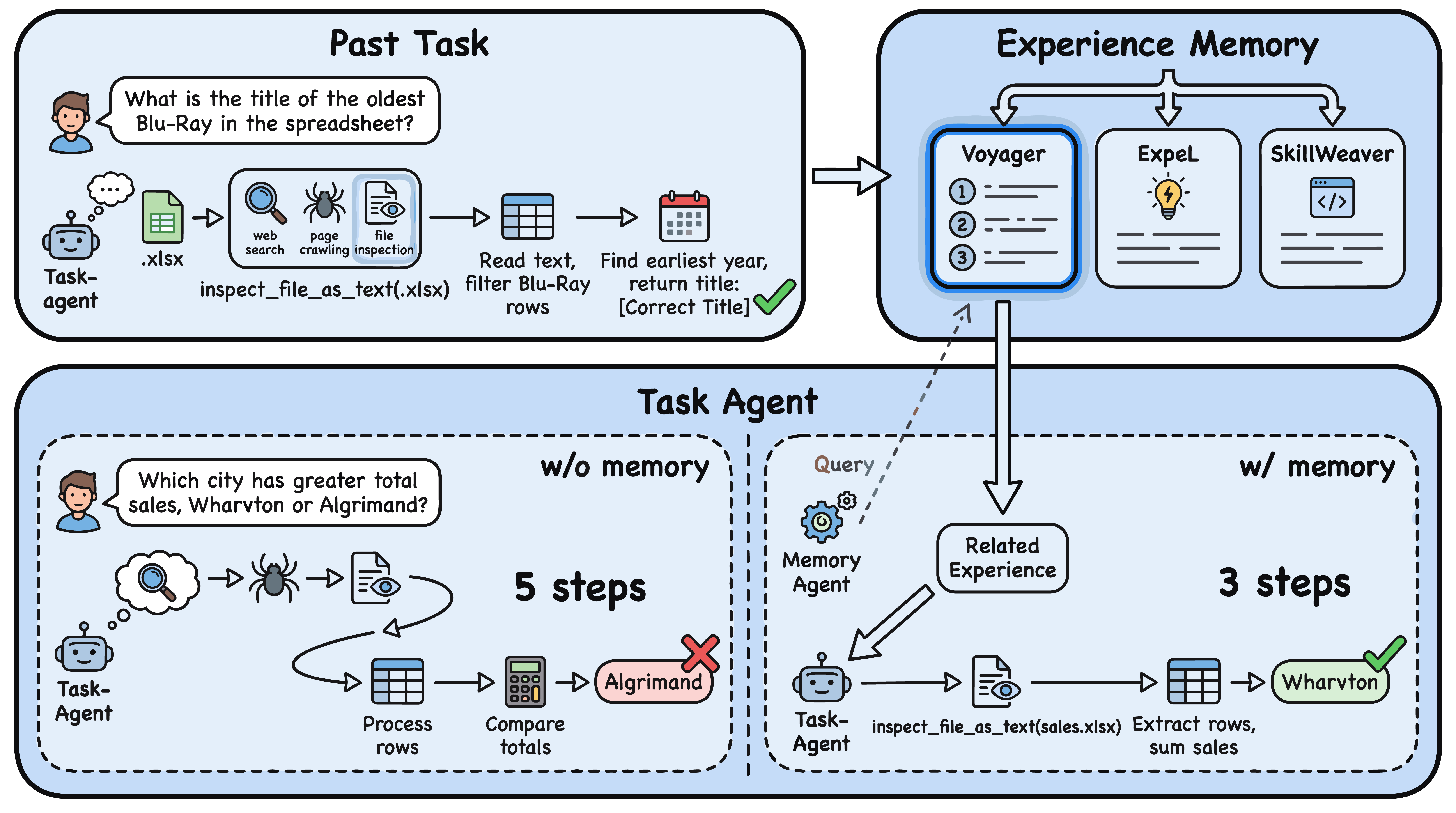}
    \caption{The task agent interacts with tools to solve tasks, while experiences from past tasks are stored in heterogeneous memory providers. By routing retrieval to the appropriate provider, the memory agent helps the task agent solve the new task correctly in fewer steps.}
    \label{fig:motivating_example}
\end{figure*}

LLM agents have rapidly extended beyond single-turn question answering into long-horizon tasks such as web research, multimodal reasoning, and tool-augmented problem solving~\cite{yao2023react,xu2025mem,shi2025look,li2025search}. Despite operating in increasingly complex environments, most agents remain largely \emph{stateless across tasks}: each invocation begins from scratch, with no mechanism to reuse prior experience or learned strategies accumulated over time. This statelessness is a fundamental obstacle to continual improvement and limits an agent's ability to amortize effort across related tasks~\cite{xu2025single,ouyang2025reasoningbank}. Memory has thus emerged as an essential mechanism for sustained agentic behavior, prompting a rich line of work on how to capture, represent, and reuse past interactions~\cite{du2025memory,long2025seeing,li2025cam,shu2026remem}.

Existing memory-augmented agents primarily differ in \emph{how} they represent past experience: summarized trajectories~\cite{wang2024voyager}, reflected insights~\cite{shinn2023reflexion}, structured knowledge~\cite{tang2025agent}, workflow templates~\cite{wang2025agent}, procedural scripts~\cite{fang2025memp}, or executable skills~\cite{zheng2025skillweaver}. Although each method reports performance gains on its target benchmark, these designs are typically proposed and evaluated in isolation. Because each approach commits to a single memory representation, a central question remains open: \emph{does any single representation remain consistently effective across different task distributions?}

To answer this question, we unify 13 representative memory methods under a common interface, evaluating each as a distinct \emph{memory provider} across three benchmarks: GAIA, WebWalkerQA, and xBench DeepSearch. Our results show strong cross-benchmark heterogeneity (see rethinking in \cref{sec:rethink}); no single method universally excels, and top performers on one benchmark frequently underperform elsewhere. Crucially, a \emph{hindsight oracle} (which considers a task solved if any provider succeeds) outperforms the best individual provider by over $18\%$. This untapped potential suggests that agent memory is less about finding a single optimal format and more about solving a routing problem over heterogeneous memory providers. Providers differ not only in the content they store but also in how they index and retrieve it, \eg, by surface similarity, structural abstraction, or composable insights. Therefore, provider choice should account for the actual retrieved content and each provider's transfer bias rather than relying only on surface-level task similarity.

Motivated by this observation, we formulate agent memory as a \emph{routing problem} over a heterogeneous provider pool (see \cref{fig:motivating_example}). For each task, a dedicated memory agent $\mathcal{A}_M$ makes three decisions across the memory lifecycle: which long-term provider to retrieve from before execution (\textsc{Begin}), whether to inject short-term working memory during execution (\textsc{In}), and which providers should store the resulting trajectory afterward (\textsc{End}). Unlike static tool routing, these decisions depend not only on provider descriptions but also on the execution context in which each decision is made.

To realize this concept, we propose MemAgent, a framework comprising two core components. The first is a \textbf{content-aware routing architecture} that spans the full memory lifecycle. At the \textsc{Begin} phase, a lightweight \emph{probe} previews each provider before routing. During the \textsc{In} phase, \emph{short-term memory gating} injects working-memory summaries only when they are likely to assist the next action. At the \textsc{End} phase, \emph{selective multi-provider storage} writes trajectories exclusively to providers where they will likely be useful, thereby enhancing future retrieval and preventing memory pollution. The second component is a \textbf{training-data synthesis pipeline}. This pipeline generates supervision for routing decisions through forced exploration and phase-specific reward signals, enabling reward-filtered supervised fine-tuning of the memory agent $\mathcal{A}_M$.

Across GAIA, WebWalkerQA, and xBench-DS, MemAgent improves average accuracy by $10.0\%$ over the baseline and beats every single-provider method on all three benchmarks. It also improves efficiency, reducing average task steps by $12\%$ while adding less than $0.3\%$ routing overhead.

In summary, our main contributions are threefold:
\begin{itemize}[leftmargin=*]
    \item We rethink agent memory through a cross-benchmark study of heterogeneous memory providers, showing that no single provider consistently dominates across tasks. Moreover, we propose a graph-structured long-term memory provider and a short-term working-memory provider.
    \item We formulate memory as a three-phase routing problem (\textsc{Begin}/\textsc{In}/\textsc{End}) and propose MemAgent, a framework comprising a content-aware routing architecture and a training-data synthesis pipeline to supervise routing decisions without ground-truth labels.
    \item Across GAIA, WebWalkerQA, and xBench-DS, MemAgent improves average accuracy by 10.0\% over the baseline and outperforms every single-provider method. Meanwhile, it reduces average task steps by $12\%$ while adding less than $0.3\%$ routing overhead.
\end{itemize}

%% file: section/related.tex
\section{Related Work}

Memory has become a central mechanism for enabling long-horizon agentic behavior, but recent work spans a broad and fragmented design space~\cite{yu2025memagent,xiangflowsearcher}. Prior surveys organize agent memory along dimensions such as functional role (\eg, working, factual, and experiential memory) and lifecycle (formation, retrieval, and refinement)~\cite{hu2025memorysurvey,zhang2025agentmemsurvey}. We group prior work into three directions.

\paragraph{Working memory and context management.}
To actively manage context windows, MEM1~\cite{zhou2026mem1} maintains a compact internal state under a fixed budget, while AgeMem~\cite{yu2026agemem} formulates memory operations as RL-trained actions. For long-horizon web and deep-research tasks, agents either reflectively reorganize their workspaces between steps (ARC~\cite{yao2026arc}, IterResearch~\cite{chen2026iterresearch}) or periodically distill trajectories into compact reasoning states (ReSum~\cite{wu2025resum}, AgentFold~\cite{ye2025agentfold}).

\paragraph{Long-term and reusable memory.}
Prior work studies cross-task memory in several forms, including summarized trajectories~\cite{wang2024voyager}, reflected successes and failures~\cite{zhao2024expel}, importance-weighted memory pools~\cite{park2023generative}, semantically indexed retrieval~\cite{zhong2024memorybank}, structured external memory~\cite{wang2025memalpha}, and utility-aware retrieval~\cite{zhang2026memrl}; MemoryAgentBench~\cite{hu2025memoryagentbench} further broadens how such systems are evaluated. A related line stores reusable procedures or skills instead of raw trajectories, as in ReMe~\cite{cao2025reme}, Memento-Skills~\cite{zhou2026mementoskills}, and MemSkill~\cite{zhang2026memskill}. These works demonstrate the value of long-term reuse, but typically retain one fixed representation.

\paragraph{Learning or evolving the memory design itself.}
Beyond improving fixed memory systems, recent works also attempt to automate memory design. For instance, ALMA~\cite{xiong2026alma} meta-learns memory designs as executable programs, whereas MemEvolve~\cite{zhang2025memevolve} evolves memory architectures across a pool of providers. Automated agent design frameworks~\cite{hu2025adas,shang2025agentsquare} treat memory components as one axis within a broader search space for agent architectures. These approaches are compatible with our observation that different memory designs exhibit complementary behavior, although they seek a better unified design for the system as a whole.

Most prior work improves a single memory system or a unified memory controller. In contrast, MemAgent treats memory heterogeneity as an opportunity: instead of proposing another memory format or globally optimized design, it learns an inference-time routing policy over heterogeneous long-term providers, together with short-term memory gating and selective multi-provider storage.

%% file: section/preliminary.tex
\section{Rethinking Agent Memory}
\label{sec:rethink}

Recent memory-augmented agents primarily differ in how they represent past experiences, utilizing formats such as experience libraries~\cite{zhao2024expel}, workflow templates~\cite{wang2025agent}, procedural scripts~\cite{fang2025memp}, knowledge graphs~\cite{zhang2025g}, and agent skills~\cite{zheng2025skillweaver}. Although each method demonstrates gains on its target benchmark, it remains unclear whether any single memory paradigm performs consistently across tasks.

\paragraph{The illusion of a universal memory.}
\label{sec:no_universal}
To test this assumption, we evaluate 13 representative memory methods, each used as the agent's sole memory system, on three diverse benchmarks: GAIA (general reasoning), WebWalkerQA (web navigation), and xBench-DS (deep search).

\begin{figure}[tbh]
    \centering
    \includegraphics[width=0.8\columnwidth]{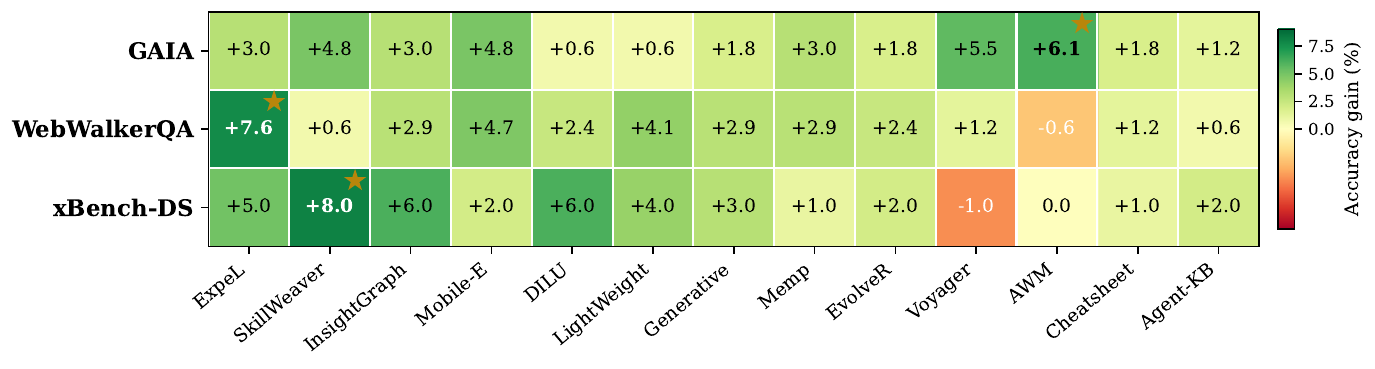}
    \vspace{-1em}
    \caption{Cross-benchmark gains of 13 memory methods over the no-memory baseline. No single method dominates across all three benchmarks. Stars mark the best method on each benchmark.}
    \label{fig:rethink_heatmap}
    \vspace{-0.5em}
\end{figure}

\cref{fig:rethink_heatmap} shows strong cross-benchmark heterogeneity. AWM leads on GAIA ($+6.1\%$) but slightly hurts WebWalkerQA ($-0.6\%$), whereas ExpeL leads on WebWalkerQA ($+7.6\%$) but is only mid-tier on GAIA. No method is best on more than one benchmark. Different tasks favor different memory representations, so simply designing a ``better'' single memory is unlikely to close the gap.

\paragraph{The routing opportunity.}
\label{sec:routing_opportunity}
Given the observed cross-benchmark complementarity, routing emerges as a promising solution. A hindsight oracle, which considers a task solved if any of the 13 single-provider systems succeeds, achieves an average accuracy of $89.6\%$ (\cref{tab:main_results}). This far exceeds the best individual provider ($71.0\%$), further supported by the fact that 10 of the 13 providers uniquely solve at least one task. These results expose a substantial opportunity for selecting among existing memory structures on a per-task basis. As illustrated in Appendix~\ref{appendix:provider_failure_analysis}, providers expose different representations and retrieval contracts; selection should therefore consider the actual retrieved content and each provider's transfer bias rather than task surface alone. This shift, however, introduces a credit-assignment challenge: task-level outcomes fail to indicate which provider or phase is responsible, because storage mechanisms receive only delayed feedback.

%% file: section/method.tex
\section{Methodology}

\begin{figure}[tb]
    \centering
    \includegraphics[width=0.8\columnwidth]{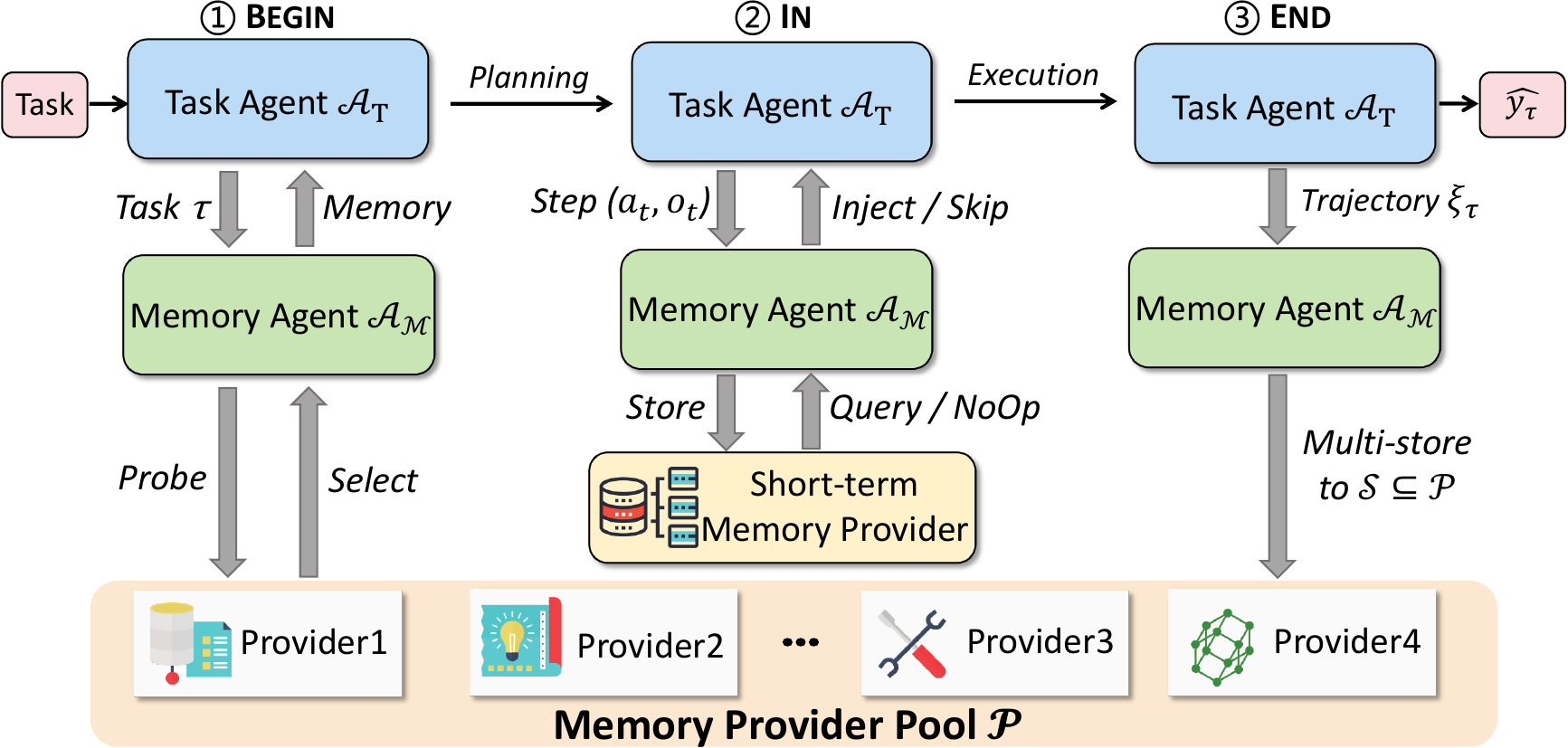}
    \caption{
   For each task, the memory agent $\mathcal{A}_M$ makes three routing decisions.
\textbf{(\romark{1}) \textsc{Begin}:} It previews all providers in
$\mathcal{P}$, selects one, and injects the retrieved long-term memory into
the task agent $\mathcal{A}_T$'s prompt.
\textbf{(\romark{2}) \textsc{In}:} At each step, the short-term memory provider
summarizes cross-step state from tool interactions $(a_t,o_t)$, and
$\mathcal{A}_M$ decides whether to inject the summary or skip.
\textbf{(\romark{3}) \textsc{End}:} After task completion, $\mathcal{A}_M$
selects a subset $\mathcal{S} \subseteq \mathcal{P}$ to store the trajectory
$\xi_\tau$ for future reuse.}
    \label{fig:inference_pipeline}
    \vspace{-1em}
\end{figure}

We introduce MemAgent, a framework in which a memory agent manages diverse memory providers for a task agent. It comprises two components: (1) a content-aware routing architecture (\cref{sec:memory_agent}), and (2) a data synthesis pipeline that provides routing supervision without ground-truth labels (\cref{sec:training_data_synthesis}). MemAgent learns a per-decision policy to route, inject, and store memories via the appropriate provider throughout task execution. \cref{fig:inference_pipeline} illustrates the overall architecture.

\subsection{Problem Formulation}
\label{sec:problem_formulation}

Consider a task agent $\mathcal{A}_T$ that solves multi-step agentic tasks. At each task $\tau$, the agent executes a trajectory $\xi_\tau = (s_0, a_0, o_0, \ldots, s_T, a_T)$ of states, actions, and observations, producing a binary outcome $y_\tau \in \{0, 1\}$ (correct or not). We augment $\mathcal{A}_T$ with a memory agent $\mathcal{A}_M$ that makes three types of memory decisions during the lifecycle of $\xi_\tau$:
(i) \textbf{\textsc{Begin}} ($d_{\text{begin}}$): before execution starts, $\mathcal{A}_M$ selects a long-term memory provider $p \in \mathcal{P}$ to retrieve relevant cross-task experience, which is injected into $\mathcal{A}_T$'s system prompt prior to $s_0$;
(ii) \textbf{\textsc{In}} ($d_{\text{in}}^{(t)}$): at each step $t$ during $\xi_\tau$, $\mathcal{A}_M$ decides whether to inject structured short-term memory or skip injection;
(iii) \textbf{\textsc{End}} ($d_{\text{end}}$): after $\xi_\tau$ completes, $\mathcal{A}_M$ selects a subset of providers $\mathcal{S} \subseteq \mathcal{P}$ to store $\xi_\tau$ for future retrieval by other tasks.

We formulate the memory agent as a policy $\mathcal{A}_M(d \mid c)$ over a decision space $\mathcal{D}$ given context $c \in \mathcal{C}$ (\eg, task state), trained to maximize the expected success rate $\max_{\mathcal{A}_M}\; \mathbb{E}_\tau[y_\tau]$.

To leverage memory diversity across varying tasks, MemAgent maintains a pool of $K$ heterogeneous memory providers. These providers share a unified interface with two standard operations: \textbf{retrieving} relevant memories and \textbf{storing} task trajectories as reusable experiences. This abstraction decouples the routing policy from provider internals, allowing $\mathcal{A}_M$ to interact with diverse memory paradigms while remaining agnostic to their underlying implementations.
In addition, we contribute a short-term memory provider for within-task working memory, and a graph-structured memory provider, InsightGraph. All providers are introduced in \cref{tab:providers} (Appendix~\ref{appendix:memory_provider_pool}).

\subsection{Content-Aware Routing}
\label{sec:memory_agent}

The memory agent $\mathcal{A}_M$ maps a phase-specific decision context $c$ to the routing actions.

\noindent\textbf{\textsc{Begin} phase: content-aware probing.} Routing based solely on the task description $q$ and static provider profiles is unreliable~\cite{fioresi2026learning}; what matters is whether a provider \emph{actually holds} relevant content. Therefore, we augment the \textsc{Begin} context $c$ with \emph{probe results}. Prior to the routing decision, $\mathcal{A}_M$ issues a lightweight parallel retrieval against all providers, obtaining a short snippet preview of each provider's most relevant memory. These previews are injected without their retrieval scores. Since providers use incompatible scoring scales, exposing scores would teach the model a spurious ``pick the highest-scored provider'' shortcut rather than forcing it to reason about semantic relevance. To mitigate positional bias, candidate providers are randomly shuffled.

\noindent\textbf{\textsc{In} phase: short-term memory gating.} To determine whether to inject short-term memory at step $t$, the context $c$ includes $q$, the step index $t$, a compact summary of the current short-term memory state, and the task agent's recent actions (\ie, recent tool calls and observations from $\xi_\tau$). This allows $\mathcal{A}_M$ to judge whether the recalled state would materially aid the next step.

\noindent\textbf{\textsc{End} phase: multi-provider storage.} To facilitate the selection of providers for storing the completed trajectory, the context $c$ comprises a trajectory summary, the binary task outcome, and provider-specific descriptions detailing what each would extract and retain.

\subsection{Training Data Synthesis}
\label{sec:training_data_synthesis}

\begin{figure}[tb]
    \centering
    \includegraphics[width=0.8\columnwidth]{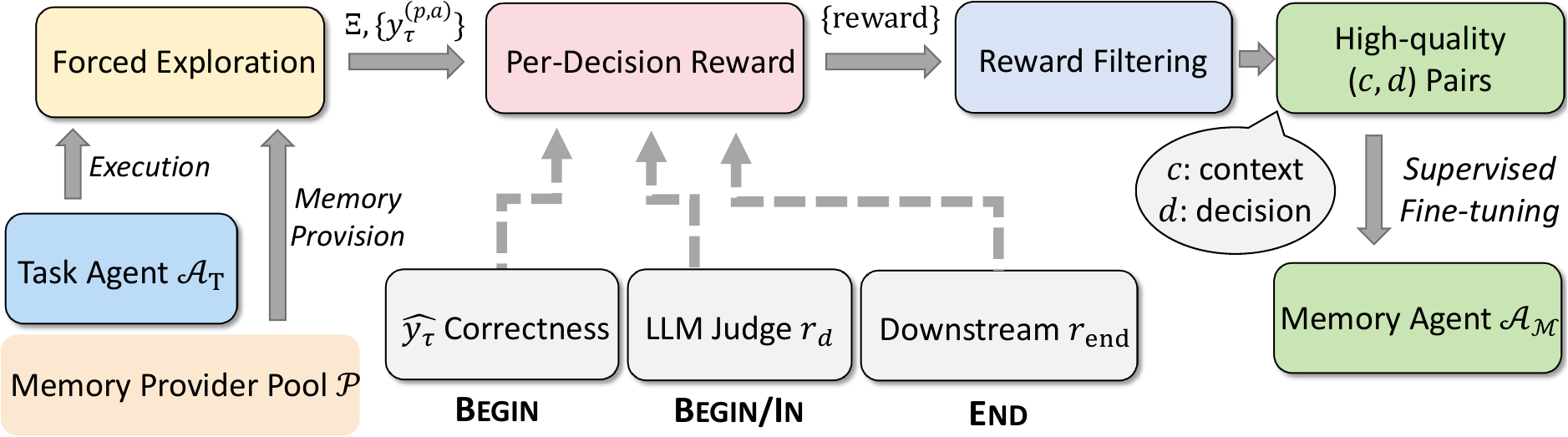}
\caption{\textbf{Training data synthesis pipeline for the memory agent.}
Forced exploration enumerates provider-action combinations for each task, producing trajectories $\Xi$ with binary outcomes $\{y_\tau^{(p,a)}\}$. Supervision signals are derived from task correctness (for \textsc{Begin}), LLM-as-Judge attribution (for \textsc{Begin} and \textsc{In}), and downstream reuse (for \textsc{End}). Finally, reward filtering retains high-quality context-decision pairs $(c,d)$ for the supervised fine-tuning of $\mathcal{A}_M$.}
    \label{fig:training_pipeline}
    \vspace{-1em}
\end{figure}

The central challenge of training $\mathcal{A}_M$ is the absence of ground-truth routing labels. We address this by synthesizing supervision through three stages: forced exploration, per-decision reward scoring, and reward-filtered data construction.

\noindent\textbf{Forced exploration.} For each training task $\tau$, we systematically iterate through all providers by forcing $\mathcal{A}_M$ to route to each $p_i \in \mathcal{P}$ in turn, while additionally varying the \textsc{In}-phase action (\textsc{Query} or \textsc{NoOp}). The task agent $\mathcal{A}_T$ executes normally in every run; only the memory routing is overridden. This process yields a trajectory set $\Xi = \{\xi_\tau^{(p,a)} : \tau \in \mathcal{T}_{\mathrm{train}},\; p \in \mathcal{P},\; a \in \mathcal{A}\}$ with corresponding recorded outcomes $\{y_\tau^{(p,a)}\}$. By observing how the \emph{same} task performs under \emph{every} provider, we better control for task difficulty and obtain comparisons across providers.

\noindent\textbf{Per-decision reward.} We derive fine-grained per-decision rewards from three complementary sources: current-task correctness, LLM-judge attribution, and downstream reuse.

\emph{Signal 1: Task correctness} (\textsc{Begin}). For task $\tau$, let $\mathcal{P}_\tau^+ = \{p : y_\tau^{(p,\cdot)} = 1\}$ denote the set of providers that led to correct outcomes. A \textsc{Begin} decision routing to $p \in \mathcal{P}_\tau^+$ receives positive reward. We further weight by discriminability: task correctness carries the strongest signal when $\mathcal{P}_\tau^+$ is a proper non-empty subset of $\mathcal{P}$, \ie, the task separates effective providers from ineffective ones.

\emph{Signal 2: LLM-as-Judge for causal attribution} (\textsc{Begin}/\textsc{In}). Task correctness provides a binary signal but cannot quantify \emph{how much} a particular retrieval or injection contributed to $y_\tau$. We employ an LLM judge~\cite{zheng2023judge} that examines each decision's context (what was retrieved or injected) alongside the task agent's subsequent actions (whether the memory was actually utilized). The judge scores multiple attribution dimensions (Appendix~\ref{appendix:judge_dimensions}), which we aggregate into a scalar reward:
\begin{equation}
    \label{eq:judge_reward}
    r_d \;=\; \big\langle \mathbf{w}^{(\phi)},\, \hat{\mathbf{s}} \big\rangle
          \;=\; \sum_{i=1}^{K_\phi} w_i^{(\phi)}\, \hat{s}_i,
    \quad
    \hat{s}_i =
    \begin{cases}
      s_i / s_{\max}, & \text{dim.\ } i\uparrow \\[-2pt]
      1 - s_i / s_{\max}, & \text{dim.\ } i\downarrow
    \end{cases},
    \quad \mathbf{w}^{(\phi)} \in \Delta^{K_\phi - 1}
\end{equation}
where $\phi \in \{\textsc{Begin}, \textsc{In}\}$ denotes the decision phase with $K_\phi$ judge dimensions, $s_i \in \{0, \dots, s_{\max}\}$ is the raw judge score, and $\uparrow/\downarrow$ indicate positive/negative dimensions (\eg, Utilization vs.\ Noise). The phase-specific weight vector $\mathbf{w}^{(\phi)} \in \Delta^{K_\phi-1}$ satisfies $w_i^{(\phi)} \geq 0$ and $\sum_i w_i^{(\phi)} = 1$.

\emph{Signal 3: Downstream reward} (\textsc{End}). For storage decisions, we define the reward based on actual downstream reuse, as a storage choice is valuable only if the reused memory \emph{also} contributes to the success of the retrieving task. Let $\xi_\tau$ be a trajectory stored in provider $p$, and let $\mathcal{T}_{>\tau}^{(p)} = \{\tau' > \tau : \textsc{Retrieved}(\tau', p) \cap \textsc{Stored}(\xi_\tau, p) \neq \emptyset\}$ denote the set of subsequent tasks where the \textsc{Begin} retrieval from $p$ contains items originating from $\xi_\tau$. We combine \emph{hit coverage} and \emph{downstream success rate} into a single scalar reward:
\begin{equation}
    \label{eq:downstream_reward}
    r_{\mathrm{end}}(\xi_\tau, p) = \tfrac{1}{2}\min\!\Big(1,\; \frac{|\mathcal{T}_{>\tau}^{(p)}|}{H}\Big) + \tfrac{1}{2} \cdot \frac{\big|\{\tau' \in \mathcal{T}_{>\tau}^{(p)} : y_{\tau'} = 1\}\big|}{\max(|\mathcal{T}_{>\tau}^{(p)}|,\, 1)}
\end{equation}
where $\textsc{Stored}(\xi_\tau, p)$ denotes the memory items written by trajectory $\xi_\tau$ to $p$, $\textsc{Retrieved}(\tau', p)$ the items retrieved by task $\tau'$ from $p$, and $H$ is a saturation constant that prevents a single heavily-reused trajectory from dominating the reward.

These three signals are complementary: task correctness defines the \emph{search space boundary} (only $y_\tau{=}1$ decisions enter the training pool), LLM-as-Judge provides \emph{quality gradation} within this space, and downstream reward provides \emph{ground-truth storage labels}.

\noindent\textbf{Reward-filtered SFT data construction.} We construct SFT data from the scored decisions using phase-specific filtering. During the \textsc{Begin} phase, task-level correctness acts as the primary filter via $\mathcal{P}_\tau^+$: (i) if $\mathcal{P}_\tau^+$ is a proper non-empty subset of $\mathcal{P}$, we keep all decisions routing to $p \in \mathcal{P}_\tau^+$; (ii) if all providers succeed ($\mathcal{P}_\tau^+ = \mathcal{P}$), we retain the top-$k$ decisions ranked by judge reward $r_d$; and (iii) if all fail ($\mathcal{P}_\tau^+ = \emptyset$), we drop the task. Decisions with $r_d = 0$, indicating empty or irrelevant retrievals, are also discarded. For the \textsc{In} phase, a threshold $r_d \geq \epsilon_{\textsc{in}}$ retains high-quality \textsc{Query} and \textsc{NoOp} decisions. Similarly, the \textsc{End} phase retains decisions where $r_{\mathrm{end}} \geq \epsilon_{\textsc{end}}$. Each training example is a (prompt, completion) pair comprising the decision context $c$ and the corresponding structured routing decision, matching the space defined in \cref{sec:memory_agent}.

\noindent\textbf{SFT training.} We fine-tune the $\mathcal{A}_M$ via LoRA~\cite{hu2022lora}, minimizing cross-entropy loss exclusively on structured decision tokens~\cite{ouyang2022instructgpt} by masking prompt tokens. To prevent data leakage from shared contexts, we perform a task-level train-validation split, ensuring all \textsc{Begin}/\textsc{In}/\textsc{End} decisions from the same task remain grouped. The end-to-end inference pipeline is summarized in Algorithm~\ref{alg:inference}.

%% file: section/experiment.tex
\section{Experiments}
In this section, we first describe our experimental setup. Then, we present our main results. We also provide ablation studies and discussions for a thorough analysis.

\subsection{Experimental Setup}

\noindent\textbf{Benchmarks.} We evaluate on three agentic benchmarks requiring multi-step tool use and reasoning: GAIA~\cite{mialon2024gaia} (across three difficulty levels), WebWalkerQA~\cite{wu2025webwalker} (multi-turn web-navigation queries), and xBench-DS~\cite{chen2025xbench} (planning and tool-use tasks). Dataset details are in Appendix~\ref{appendix:dataset_details}.

\noindent\textbf{Agent framework.} We use Flash-Searcher~\cite{qin2026flashsearcher}, a single-agent deep research framework as the task agent, equipped with four tools: \texttt{WebSearch} (search engine queries), \texttt{CrawlPage} (URL content extraction), \texttt{VisualInspector} (image analysis), and \texttt{AudioInspector} (audio transcription). The task agent is powered by GPT-5-mini via API.
The memory agent is Qwen3.5-4B~\cite{qwen35blog} fine-tuned with LoRA ($r{=}16$, $\alpha{=}32$, dropout$=0.05$; 0.84\% trainable).

\noindent\textbf{Training configuration.} Following MemEvolve~\cite{zhang2025memevolve}, training uses 53 GAIA Level-1 tasks and 120 synthetically generated multi-hop tasks from TaskCraft~\cite{shi2025taskcraft}. Per-decision reward scoring uses GPT-5-mini as the LLM-as-Judge.
SFT training uses learning rate $2{\times}10^{-5}$, cosine schedule, warmup ratio 0.1, effective batch size 64, and maximum sequence length 4,096.

\noindent\textbf{Evaluation protocol.} Storage accumulated during forced exploration is seeded into evaluation runs, ensuring that the memory agent has access to a memory pool at inference time. Evaluation follows an online protocol, where the experiential memory base is updated on-the-fly as the agent system processes a continuous stream of tasks. The task agent is capped at 15 steps per task on GAIA and 40 steps on WebWalkerQA and xBench-DS, applied uniformly across all methods.

\noindent\textbf{Baselines.}
We compare MemAgent against different memory methods. (i) \textbf{No Memory} runs the task agent without any memory augmentation. (ii) \textbf{Random Routing} replaces the memory agent with random routing decisions. (iii) \textbf{MemAgent w/o training} directly employs the Qwen3.5-4B model. (iv) \textbf{Individual Memory Methods} evaluate each of the 13 long-term memory providers in isolation. (v) \textbf{PromptRoute} replaces the SFT-trained memory agent with GPT-5-mini, utilizing a prompt that includes per-provider statistics, routing rules, and representative cases mined from forced-exploration trajectories (see Appendix~\ref{appendix:promptroute_dkb}). (vi) \textbf{MemAgent (5 providers)} restricts our agent to a curated subset (ExpeL, AWM, SkillWeaver, Voyager, and InsightGraph) selected for cross-benchmark complementarity, thereby isolating the effect of the provider pool size. (vii) \textbf{Oracle} chooses the successful provider for each task when available, defining the upper bound for routing. (viii) \textbf{All-Concat} retrieves from all memory providers at the \textsc{Begin} phase and concatenates their outputs. (ix) \textbf{All-Fuse} performs the same memory retrieval but uses an LLM to consolidate the retrieved content into a single memory note.

\subsection{Main Results}

\begin{table*}[tbp]
\centering
\caption{Accuracy (\%) on GAIA, WebWalkerQA, and xBench-DS. Gray deltas show percentage-point changes vs. the no-memory baseline. GAIA results are detailed by difficulty levels ($L1$--$L3$). \textbf{Bold} and \underline{underline} denote the best and second-best memory methods.}
\label{tab:main_results}
\resizebox{\textwidth}{!}{%
\begin{tabular}{l cccc c c c}
\toprule
\multicolumn{1}{c}{\multirow{2}{*}{\textbf{Method}}} & \multicolumn{4}{c}{\textbf{GAIA}} & \multirow{2}{*}{\textbf{WebWalkerQA}} & \multirow{2}{*}{\textbf{xBench-DS}} & \multirow{2}{*}{\textbf{Avg.}} \\
\cmidrule(lr){2-5}
 & Overall & Level 1 & Level 2 & Level 3 &  &  &  \\
\midrule
No Memory & 58.8 & 73.6 & 54.7 & 42.3 & 69.4 & 69.0 & 65.7 \\
\midrule
\multicolumn{8}{l}{\textit{Individual memory methods}} \\
Agent-KB~\cite{tang2025agent}              & 60.0 {\scriptsize\color{gray}+1.2} & 73.6 & 54.7 & 50.0 & 70.0 {\scriptsize\color{gray}+0.6} & 71.0 {\scriptsize\color{gray}+2.0} & 67.0 {\scriptsize\color{gray}+1.3} \\
LightWeight~\cite{zhang2025memevolve}           & 59.4 {\scriptsize\color{gray}+0.6} & 67.9 & 61.6 & 34.6 & 73.5 {\scriptsize\color{gray}+4.1} & 73.0 {\scriptsize\color{gray}+4.0} & 68.6 {\scriptsize\color{gray}+2.9} \\
ExpeL~\cite{zhao2024expel}       & 61.8 {\scriptsize\color{gray}+3.0} & 73.6 & 61.6 & 38.5 & \underline{77.1} {\scriptsize\color{gray}+7.6} & 74.0 {\scriptsize\color{gray}+5.0} & \underline{71.0} {\scriptsize\color{gray}+5.2} \\
SkillWeaver~\cite{zheng2025skillweaver}           & 63.6 {\scriptsize\color{gray}+4.8} & 71.7 & 66.3 & 38.5 & 70.0 {\scriptsize\color{gray}+0.6} & \underline{77.0} {\scriptsize\color{gray}+8.0} & 70.2 {\scriptsize\color{gray}+4.5} \\
Voyager~\cite{wang2024voyager}               & 64.2 {\scriptsize\color{gray}+5.5} & 77.4 & 64.0 & 38.5 & 70.6 {\scriptsize\color{gray}+1.2} & 68.0 {\scriptsize\color{gray}$-$1.0} & 67.6 {\scriptsize\color{gray}+1.9} \\
InsightGraph~(Ours)        & 61.8 {\scriptsize\color{gray}+3.0} & 69.8 & 59.3 & 53.8 & 72.4 {\scriptsize\color{gray}+2.9} & 75.0 {\scriptsize\color{gray}+6.0} & 69.7 {\scriptsize\color{gray}+4.0} \\
DILU~\cite{wen2024dilu}                  & 59.4 {\scriptsize\color{gray}+0.6} & 69.8 & 61.6 & 30.8 & 71.8 {\scriptsize\color{gray}+2.4} & 75.0 {\scriptsize\color{gray}+6.0} & 68.7 {\scriptsize\color{gray}+3.0} \\
Generative~\cite{park2023generative}            & 60.6 {\scriptsize\color{gray}+1.8} & 75.5 & 58.1 & 38.5 & 72.4 {\scriptsize\color{gray}+2.9} & 72.0 {\scriptsize\color{gray}+3.0} & 68.3 {\scriptsize\color{gray}+2.6} \\
Memp~\cite{fang2025memp}                  & 61.8 {\scriptsize\color{gray}+3.0} & 73.6 & 61.6 & 38.5 & 72.4 {\scriptsize\color{gray}+2.9} & 70.0 {\scriptsize\color{gray}+1.0} & 68.1 {\scriptsize\color{gray}+2.3} \\
Cheatsheet~\cite{suzgun2026dynamic}            & 60.6 {\scriptsize\color{gray}+1.8} & 69.8 & 59.3 & 46.2 & 70.6 {\scriptsize\color{gray}+1.2} & 70.0 {\scriptsize\color{gray}+1.0} & 67.1 {\scriptsize\color{gray}+1.4} \\
AWM~\cite{wang2025agent}    & \underline{64.8} {\scriptsize\color{gray}+6.1} & 75.5 & \underline{64.0} & 46.2 & 68.8 {\scriptsize\color{gray}$-$0.6} & 69.0 {\scriptsize\color{gray}+0.0} & 67.5 {\scriptsize\color{gray}+1.8} \\
EvolveR~\cite{wu2025evolver}               & 60.6 {\scriptsize\color{gray}+1.8} & 66.0 & 65.1 & 34.6 & 71.8 {\scriptsize\color{gray}+2.4} & 71.0 {\scriptsize\color{gray}+2.0} & 67.8 {\scriptsize\color{gray}+2.1} \\
Mobile-E~\cite{wang2025mobile}              & 63.6 {\scriptsize\color{gray}+4.8} & \underline{81.1} & 62.8 & 30.8 & 74.1 {\scriptsize\color{gray}+4.7} & 71.0 {\scriptsize\color{gray}+2.0} & 69.6 {\scriptsize\color{gray}+3.9} \\
\midrule
\multicolumn{8}{l}{\textit{Routing methods}} \\
Random Routing   & 58.2{\scriptsize\color{gray}$-$0.6} & 67.9 & 58.1 & 38.5 & 72.4 {\scriptsize\color{gray}+3.0} & 71.0 {\scriptsize\color{gray}+2.0} & 67.2 {\scriptsize\color{gray}+1.5} \\
MemAgent w/o training & 63.6 {\scriptsize\color{gray}+4.8} & 71.7 & 67.4 & 34.6 & 70.6 {\scriptsize\color{gray}+1.2} & 70.0 {\scriptsize\color{gray}+1.0} & 68.1 {\scriptsize\color{gray}+2.4} \\
MemAgent (5 providers) & 63.6 {\scriptsize\color{gray}+4.8} & 77.4 & 61.6 & 42.3 & 72.9 {\scriptsize\color{gray}+3.5} & 73.0 {\scriptsize\color{gray}+4.0} & 69.8 {\scriptsize\color{gray}+4.1} \\
PromptRoute   & 65.5 {\scriptsize\color{gray}+6.7} & 77.4 & 64.0 & 46.2 & 71.2 {\scriptsize\color{gray}+1.8} & 76.0 {\scriptsize\color{gray}+7.0} & 70.9 {\scriptsize\color{gray}+5.2} \\
All-Concat      & 62.4 {\scriptsize\color{gray}+3.6} & 79.2 & 61.6 & 30.8 & 74.1 {\scriptsize\color{gray}+4.7} & 73.0 {\scriptsize\color{gray}+4.0} & 69.8 {\scriptsize\color{gray}+4.1} \\
All-Fuse        & 63.6 {\scriptsize\color{gray}+4.8} & 83.0 & 60.5 & 34.6 & 75.3 {\scriptsize\color{gray}+5.9} & 74.0 {\scriptsize\color{gray}+5.0} & \underline{71.0} {\scriptsize\color{gray}+5.3} \\
\textbf{MemAgent (Ours)} & \textbf{69.7} {\scriptsize\color{gray}+10.9} & \textbf{83.0} & \textbf{67.4} & \textbf{50.0} & \textbf{79.4} {\scriptsize\color{gray}+10.0} & \textbf{78.0} {\scriptsize\color{gray}+9.0} & \textbf{75.7} {\scriptsize\color{gray}+10.0} \\
\midrule
\multicolumn{8}{l}{\textit{Upper bound}} \\
Oracle (13 providers) & 87.3 {\scriptsize\color{gray}+28.5} & 94.3 & 87.2 & 73.1 & 89.4 {\scriptsize\color{gray}+20.0} & 92.0 {\scriptsize\color{gray}+23.0} & 89.6 {\scriptsize\color{gray}+23.8} \\
\bottomrule
\end{tabular}%
}
\vspace{-0.5em}
\end{table*}

\cref{tab:main_results} presents the results across all three benchmarks. MemAgent achieves consistent gains over the no-memory baseline, averaging a 10.0\% improvement overall. Crucially, MemAgent surpasses every single-provider baseline on all benchmarks: +4.9\% on GAIA (vs.\ AWM), +2.3\% on WebWalkerQA (vs.\ ExpeL), and +1.0\% on xBench (vs.\ SkillWeaver). While different individual methods dominate specific benchmarks, MemAgent consistently outperforms them all. This validates that learned routing effectively captures complementary strengths across the provider pool. On GAIA, these improvements remain consistent across all difficulty levels. To further disentangle the sources of these gains, we examine two controlled variants that isolate the roles of routing and of training. Replacing the memory agent with a random policy (Random Routing) fails to recover MemAgent's performance, indicating that the improvements arise from \emph{learned} rather than arbitrary routing. MemAgent w/o training likewise falls short of MemAgent, demonstrating that the model's instruction-following ability alone is insufficient and that training is essential for effective routing.

\noindent\textbf{PromptRoute.} We investigate whether routing knowledge can be distilled into a prompt instead of learned via SFT. PromptRoute replaces the SFT-trained agent with a frozen LLM, optimizing its prompt using a Decision Knowledge Base (DKB). This DKB extracts performance statistics, routing rules, and representative examples from forced exploration trajectories for inference-time injection. PromptRoute averages 70.9\% across benchmarks (+5.2\% over the no-memory baseline), proving the efficacy of LLM-mined knowledge. However, MemAgent still outperforms it, confirming that SFT captures finer-grained routing distinctions than prompt-based retrieval alone. We view LLM-mined prompt optimization as a promising direction for future work.

\noindent\textbf{Provider pool size.} MemAgent (5 providers) restricts routing to a curated subset of five complementary providers. This variant achieves 69.8\% overall, trailing the full 13-provider system. Although it outperforms the no-memory baseline and rivals most single-provider methods, this gap underscores the importance of provider diversity. Excluded providers (\eg, LightWeight, Memp, DILU, and Mobile-E) evidently offer complementary strengths that the full routing model successfully exploits.

\begin{table*}[tbp]
\centering
\begin{minipage}[t]{0.51\textwidth}
\centering
\caption{Training-pipeline ablation on GAIA. Accuracy (\%); $\Delta$ = drop relative to the full model.}
\vspace{-0.5em}
\label{tab:train_ablation}
\vspace{1pt}
\scriptsize
\setlength{\tabcolsep}{2.6pt}
\renewcommand{\arraystretch}{1.04}
\begin{tabular}{@{}p{0.54\linewidth}ccccc@{}}
\toprule
\multicolumn{1}{c}{\multirow{2}{*}{\textbf{Configuration}}} & \multicolumn{4}{c}{\textbf{Accuracy (\%)}} & \multirow{2}{*}{\textbf{$\Delta$}} \\
\cmidrule(lr){2-5}
 & \textbf{Ov.} & \textbf{L1} & \textbf{L2} & \textbf{L3} & \\
\midrule
Base (Judge all, no probe, no filtering) & 60.6 & 75.5 & 61.6 & 26.9 & {\color{gray}--} \\
w/o \textsc{Begin} correctness filtering & 62.4 & 79.2 & 60.5 & 34.6 & +7.3 \\
w/o Probe score removal & 63.0 & 77.4 & 64.0 & 30.8 & +6.7 \\
w/o \textsc{In} threshold filtering & 63.6 & 79.2 & 61.6 & 38.5 & +6.1 \\
w/o \textsc{End} downstream reward & 66.7 & 75.5 & \textbf{68.6} & 42.3 & +3.0 \\
\midrule
\textbf{MemAgent (Full)} & \textbf{69.7} & \textbf{83.0} & 67.4 & \textbf{50.0} & {\color{gray}--} \\
\bottomrule
\end{tabular}
\end{minipage}
\hfill
\begin{minipage}[t]{0.46\textwidth}
\centering
\caption{Inference-architecture ablation on GAIA. Accuracy (\%); selective stores to chosen providers, all-store to all 13 providers.}
\label{tab:arch_ablation}
\vspace{1pt}
\scriptsize
\setlength{\tabcolsep}{3.2pt}
\renewcommand{\arraystretch}{1.04}
\begin{tabular}{@{}ccccccc@{}}
\toprule
\multirow{2}{*}{\textbf{Begin}} & \multirow{2}{*}{\textbf{In}} & \multirow{2}{*}{\textbf{End}} & \multicolumn{4}{c}{\textbf{Accuracy (\%)}} \\
\cmidrule(lr){4-7}
 &  &  & \textbf{Ov.} & \textbf{L1} & \textbf{L2} & \textbf{L3} \\
\midrule
\ding{55} & \ding{55} & \ding{55} & 58.8 & 73.6 & 54.7 & 42.3 \\
\ding{55} & \ding{51} & \ding{55} & 63.0 & 77.4 & 64.0 & 30.8 \\
\ding{51} & \ding{51} & All-store & 65.5 & 81.1 & 64.0 & 38.5 \\
\ding{51} & \ding{51} & \textbf{Selective} & \textbf{69.7} & \textbf{83.0} & \textbf{67.4} & \textbf{50.0} \\
\bottomrule
\end{tabular}
\end{minipage}
\vspace{-0.5em}
\end{table*}

\noindent\textbf{Retrieving from all providers.}
Retrieving and combining memories from every provider can introduce redundant, irrelevant, or conflicting context. The all-provider baselines in \cref{tab:main_results} test whether broad coverage alone can replace selective routing, and support selecting provider-specific content rather than indiscriminate aggregation.

\subsection{Analysis and Discussion}
\label{sec:analysis}

\paragraph{Ablation Studies.}
\label{sec:ablations}

We conduct two systematic ablation studies to isolate the contributions of the training data synthesis pipeline and the three-phase inference architecture.
\cref{tab:train_ablation} presents a leave-one-out ablation on GAIA. Compared to a \textit{Base} configuration using heuristic LLM-as-Judge rewards (without probing or filtering), our four components yield a combined +9.1\% improvement. \textsc{Begin} correctness filtering is the most impactful (+7.3\%); using ground-truth outcomes rather than proxy scores is critical for harder tasks (boosting L3 from 34.6\% to 50.0\%). Removing probe scores (+6.7\%) prevents the agent from exploiting incompatible metric scales (\eg, raw counts vs. cosine similarities) to naively select the highest-scoring provider. \textsc{In}-phase threshold filtering curbs short-term memory over-querying, which otherwise degrades long-horizon performance. Finally, the downstream \textsc{End} reward (+3.0\%) optimizes storage decisions by measuring actual memory reuse.

\cref{tab:arch_ablation} ablates the three-phase inference by varying whether long-term memory is retrieved, whether short-term memory is injected, and whether storage is selective or indiscriminate. The \textsc{In}-only configuration (rule-based short-term memory injection without long-term retrieval or storage) achieves a 4.2\,\% gain over the baseline, confirming the value of within-task working memory. The force-all-store ablation, which stores every trajectory across all providers, achieves only 65.5\%. This occurs because indiscriminate storage pollutes the retrieval pools with low-relevance memories. This highlights the necessity of selective \textsc{End}-phase routing; without it, the noise from stored low-quality memories offsets the benefits of long-term memory.

\noindent
\begin{minipage}[t]{0.52\textwidth}
\vspace{0pt}
\paragraph{Task-Agent Transfer.}
We replace GPT-5-mini with GPT-5.6 Luna and transfer the memory agent without retraining. \cref{tab:task_agent_transfer} shows that MemAgent outperforms both the no-memory setting and the best fixed provider overall and at every difficulty level, indicating that learned routing remains effective across task-agent backbones. Appendix~\ref{appendix:router_capacity} further examines the effect of router capacity.
\end{minipage}
\hfill
\begin{minipage}[t]{0.45\textwidth}
\vspace{0pt}
\centering
\captionsetup{hypcap=false}
\captionof{table}{Zero-shot transfer to GPT-5.6 Luna on GAIA.}
\label{tab:task_agent_transfer}
\scriptsize
\setlength{\tabcolsep}{2.8pt}
\resizebox{\linewidth}{!}{%
\begin{tabular}{lcccc}
\toprule
\textbf{Method} & \textbf{Overall} & \textbf{L1} & \textbf{L2} & \textbf{L3} \\
\midrule
No Memory             & 67.9 & 77.4 & 68.6 & 46.2 \\
AWM (best provider)    & 69.7 & 77.4 & 68.6 & 57.7 \\
MemAgent (zero-shot)  & \textbf{71.5} & \textbf{79.2} & \textbf{69.8} & \textbf{61.5} \\
\bottomrule
\end{tabular}%
}
\end{minipage}

\paragraph{Inference Overhead.}

A natural concern is the computational cost of memory-augmented inference; \cref{tab:overhead} details this per-task overhead on GAIA. The volume of memory text injected into the task agent's context is minimal, totaling $\sim$1.2K tokens per task (less than 0.5\% of the task agent's average tokens per task). While injection overhead is inherent to all memory-augmented approaches, MemAgent's injection size remains moderate compared to 13 other providers, such as ExpeL at $\sim$970 tokens (see Appendix~\ref{appendix:injection_overhead}). MemAgent's unique routing cost involves parallel probing and inference via a 4B-parameter model. This step adds only 1.2 seconds per task, or less than 0.3\% of average latency. Crucially, memory augmentation actually \emph{reduces} the overall number of task steps. MemAgent requires an average of 8.0 steps compared to 9.1 for the no-memory baseline (a 12\% reduction). This efficiency gain is most pronounced on harder tasks, where L3 steps drop from 12.3 to 10.7, as retrieved memories help the agent avoid unproductive exploration.

\begin{table}[t]
\centering
\caption{\textbf{Inference overhead of MemAgent on GAIA.} Injection sizes count memory tokens added to each task-agent prompt. Routing latency is MemAgent's extra wall-clock cost per task.}
\label{tab:overhead}
\small
\setlength{\tabcolsep}{4pt}
\renewcommand{\arraystretch}{0.95}
\begin{tabular}{lcc}
\toprule
\textbf{Component} & \textbf{Per-task cost} & \textbf{Notes} \\
\midrule
\multicolumn{3}{l}{\textit{Memory injection (tokens inserted into context):}} \\
\quad \textsc{Begin} (long-term retrieval)  & $\sim$370 tokens  & median 268, max 2.2K \\
\quad \textsc{In} (short-term, per step)    & $\sim$210 tokens  & 3.8 steps/task triggered \\
\quad Total injected per task               & $\sim$1.2K tokens & $<$0.5\% of task tokens \\
\midrule
\multicolumn{3}{l}{\textit{Routing cost (MemAgent-specific):}} \\
\quad Probe (13 providers, parallel)        & $\sim$75\,ms   & one-time at \textsc{Begin} \\
\quad Memory agent inference (4B)           & $\sim$80\,ms/decision & $\sim$9 decisions/task \\
\quad Total routing latency                 & $\sim$1.2\,s/task & $<$0.3\% of task time \\
\midrule
\multicolumn{3}{l}{\textit{Task efficiency:}} \\
\quad Avg.\ steps (No Memory $\to$ MemAgent)  & 9.1 $\to$ 8.0 steps  & $-$12\% \\
\bottomrule
\end{tabular}
\vspace{-0.8em}
\end{table}

%% file: section/conclusion.tex
\section{Conclusion}

In this paper, we revisit agent memory from the perspective of heterogeneity and introduce MemAgent, a three-phase routing framework. Its content-aware architecture and data-synthesis pipeline supervise long-term retrieval, short-term memory use, and selective storage across providers. Across three benchmarks, MemAgent improves average accuracy by 10.0\% and outperforms single-provider baselines while reducing task steps with minimal routing overhead. Overall, our results support adaptive, content-aware orchestration throughout the memory lifecycle.

%% file: section/appendix.tex
\section*{Appendix Contents}
\begingroup
\small
\renewcommand{\arraystretch}{1.12}
\begin{tabularx}{\textwidth}{@{}p{0.09\textwidth}Xr@{}}
\textbf{A} & \hyperref[appendix:reproducibility]{Implementation Details} & \pageref{appendix:reproducibility} \\
 & \hspace{1em}\hyperref[appendix:memory_provider_pool]{A.1 Memory Provider Pool} & \pageref{appendix:memory_provider_pool} \\
 & \hspace{1em}\hyperref[appendix:dataset_details]{A.2 Dataset Details} & \pageref{appendix:dataset_details} \\
 & \hspace{1em}\hyperref[appendix:training_details]{A.3 Training Details} & \pageref{appendix:training_details} \\
 & \hspace{1em}\hyperref[appendix:eval_details]{A.4 Evaluation Details} & \pageref{appendix:eval_details} \\
 & \hspace{1em}\hyperref[appendix:provider_params]{A.5 Provider Implementation Parameters} & \pageref{appendix:provider_params} \\[2pt]
\textbf{B} & \hyperref[appendix:method_details]{Method Details} & \pageref{appendix:method_details} \\
 & \hspace{1em}\hyperref[appendix:inference_pipeline]{B.1 Inference Pipeline} & \pageref{appendix:inference_pipeline} \\
 & \hspace{1em}\hyperref[appendix:judge_dimensions]{B.2 LLM-as-Judge Reward Design} & \pageref{appendix:judge_dimensions} \\
 & \hspace{1em}\hyperref[appendix:insight_graph]{B.3 InsightGraph} & \pageref{appendix:insight_graph} \\
 & \hspace{1em}\hyperref[appendix:short_term_memory]{B.4 Short-Term Memory Provider} & \pageref{appendix:short_term_memory} \\
 & \hspace{1em}\hyperref[appendix:prompt_templates]{B.5 Decision Prompt Format and Output Schema} & \pageref{appendix:prompt_templates} \\[2pt]
\textbf{C} & \hyperref[appendix:additional_analyses]{Further Analyses} & \pageref{appendix:additional_analyses} \\
 & \hspace{1em}\hyperref[appendix:router_capacity]{C.1 Router Capacity} & \pageref{appendix:router_capacity} \\
 & \hspace{1em}\hyperref[appendix:promptroute_dkb]{C.2 PromptRoute: Decision Knowledge Base} & \pageref{appendix:promptroute_dkb} \\
 & \hspace{1em}\hyperref[appendix:routing_behavior]{C.3 Routing Behavior} & \pageref{appendix:routing_behavior} \\
 & \hspace{1em}\hyperref[appendix:injection_overhead]{C.4 Injection Overhead} & \pageref{appendix:injection_overhead} \\[2pt]
\textbf{D} & \hyperref[appendix:provider_failure_analysis]{Why Do Memory Providers Exhibit Complementary Behavior?} & \pageref{appendix:provider_failure_analysis} \\
 & \hspace{1em}\hyperref[appendix:systematic_failures]{D.1 Three Representational Families and Their Failure Modes} & \pageref{appendix:systematic_failures} \\
 & \hspace{1em}\hyperref[appendix:theoretical_lens]{D.2 A Qualitative Lens: Task Surface and Transfer Requirements} & \pageref{appendix:theoretical_lens} \\
 & \hspace{1em}\hyperref[sec:bridges]{D.3 Case Studies: Complementary Transfer Patterns in Storage} & \pageref{sec:bridges} \\
 & \hspace{1em}\hyperref[appendix:implications]{D.4 Implications} & \pageref{appendix:implications} \\
\end{tabularx}
\endgroup

\clearpage

\section{Implementation Details}
\label{appendix:reproducibility}

This section outlines the memory provider pool, dataset, training, and evaluation details for our experiments, followed by the implementation parameters for each provider.

\subsection{Memory Provider Pool}
\label{appendix:memory_provider_pool}

Table~\ref{tab:providers} summarizes the comprehensive provider pool utilized by MemAgent. This pool comprises 13 heterogeneous long-term providers spanning exploration memory, experience learning, workflow induction, procedural memory, principle evolution, and graph-structured memory, alongside a short-term memory provider designed for within-task state aggregation. Notably, we introduce two custom providers to this framework: InsightGraph, a graph-structured long-term provider, and a rule-based short-term memory provider. Comprehensive implementation details for these novel components are available in Appendix~\ref{appendix:insight_graph} and Appendix~\ref{appendix:short_term_memory}.

\begin{table}[htbp]
\centering
\caption{\textbf{Memory providers in MemAgent.} The pool comprises 13 heterogeneous long-term providers spanning diverse memory paradigms, plus a short-term memory provider for within-task state aggregation. Storage Strategy: the specific information extracted and saved from past trajectories; Retrieval Mechanism: the method used to query relevant memories for new tasks.}
\label{tab:providers}
\resizebox{\textwidth}{!}{%
\begin{tabular}{llp{4.5cm}p{4.5cm}}
\toprule
\textbf{Provider} & \textbf{Memory Paradigm} & \textbf{Storage Strategy} & \textbf{Retrieval Mechanism} \\
\midrule

Voyager~\cite{wang2024voyager} & Exploration memory & Refined trajectory summaries with embeddings & Semantic search over task similarity \\
ExpeL~\cite{zhao2024expel} & Experience learning & Dual-path: insights from all outcomes + refined trajectories from successes & Hybrid insight + trajectory retrieval \\
Generative~\cite{park2023generative} & Scored memory pool & Task queries + trajectory summaries & Vector retrieval + LLM importance scoring \\
DILU~\cite{wen2024dilu} & Query-trajectory mapping & Query embeddings + trajectory summaries & Semantic search on query similarity \\
AWM~\cite{wang2025agent} & Workflow templates (SOPs) & LLM-induced abstract workflow patterns & Vector search over workflow templates \\
Mobile-E~\cite{wang2025mobile} & Keyword-indexed tips & Tips (text) + Shortcuts (structured sequences) & Keyword-based retrieval \\
Cheatsheet~\cite{suzgun2026dynamic} & Global best practices & Continuously updated cheatsheet of rules & LLM-curated synthesis \\
SkillWeaver~\cite{zheng2025skillweaver} & Procedural code memory & Trajectory $\rightarrow$ generalized Python functions & Skill matching $\rightarrow$ dynamic tool wrapping \\
Agent-KB~\cite{tang2025agent} & Structured knowledge base & Full workflow decomposition into planning \& experience fields & Student/Teacher guidance injection \\
Memp~\cite{fang2025memp} & Procedural scripts & Process script induction / self-correction & Script similarity matching \\
EvolveR~\cite{wu2025evolver} & Evolving principles & Principle extraction + semantic merge/prune & Principle retrieval with examples \\
LightWeight~\cite{zhang2025memevolve} & Dual-layer factual memory & Strategic (planning) + operational (tool usage) extraction & Semantic search over both layers \\
\midrule
InsightGraph~(Ours) & Knowledge graph & LLM-extracted insight nodes with semantic edges & Hybrid search + graph expansion \\
Short-Term Memory~(Ours) & Structured working memory & Rule-based extraction of facts, URLs, searches & Template-based sectioned output \\
\bottomrule
\end{tabular}%
}
\end{table}

\subsection{Dataset Details}
\label{appendix:dataset_details}

The training set comprises 173 tasks: 53 GAIA Level-1 tasks and 120 synthetic multi-hop tasks from TaskCraft~\cite{shi2025taskcraft}. Each TaskCraft task is a two-hop question covering domains such as academic research, financial reports, and scientific papers. These tasks require tool combinations like (web search, web search) or (image analysis, web search)~\cite{liu2024agentbench,wu2026webdancer}. An example is as follows:

\begin{quote}
\textit{``According to the January 9, 2025 special feature on Artsy, what did Clare Lilley forecast regarding diversity and craft in the art world?''} \hfill [web $\to$ web]
\end{quote}

Following MemEvolve~\cite{zhang2025memevolve}, the 53 GAIA L1 tasks are used both to populate the memory providers during forced exploration and to evaluate downstream reuse. We therefore treat GAIA L1 as an experience-reuse setting rather than a strict held-out generalization test. The router itself is supervised only on provider selection, not on gold answers, and this protocol is intended to mimic deployment scenarios in which the memory pool has already been populated through prior interactions. By contrast, the remaining 112 GAIA L2/L3 tasks and all WebWalkerQA and xBench-DS tasks are disjoint from training and serve as held-out generalization benchmarks.

\subsection{Training Details}
\label{appendix:training_details}

We instantiate the memory agent with Qwen3.5-4B~\cite{qwen35blog} and fine-tune it with LoRA ($r{=}16$, $\alpha{=}32$, dropout$=0.05$) applied to all attention and MLP projections, yielding 0.84\% trainable parameters. SFT minimizes cross-entropy loss on completion tokens only (prompt tokens are masked):
\begin{equation}
    \mathcal{L}_{\text{SFT}} = -\frac{1}{N} \sum_{i=1}^{N} \log P_\theta(y_i \mid x_i)
\end{equation}
where $y_i$ are the completion tokens and $x_i$ the corresponding input contexts. Training uses learning rate $2{\times}10^{-5}$ with cosine annealing and 10\% linear warmup, weight decay 0.01, effective batch size 64, maximum sequence length 4,096 tokens. We train for 3 epochs and select the final checkpoint via validation on a held-out subset of decisions. The downstream reward (\cref{eq:downstream_reward}) uses saturation constant $H=5$, and the \textsc{End}-phase SFT filter uses threshold $\epsilon_{\textsc{end}}=0.5$.

\subsection{Evaluation Details}
\label{appendix:eval_details}

\paragraph{Task agent and tools.}
The task agent is a single-agent deep research framework (Flash-Searcher~\cite{qin2026flashsearcher}) powered by GPT-5-mini via API, equipped with four tools: \texttt{WebSearch}, \texttt{CrawlPage}, \texttt{VisualInspector}, and \texttt{AudioInspector}. At each step the agent either issues one tool call or emits a final answer.
Within each benchmark, every method (no-memory, individual providers, random routing, PromptRoute, MemAgent) uses the same budget. If the step budget is reached before a final answer, the agent is prompted to answer from its context; the same correctness judge evaluates this output.

\paragraph{Correctness judging.}
For GAIA and WebWalkerQA, GPT-5-mini determines correctness by comparing the agent's final answer against the ground truth. This comparison relies on a strict equivalence rubric, requiring the exact same entity, an identical numeric value (regardless of formatting), or a semantically equivalent short phrase. For xBench-DS, we follow the benchmark's original protocol, employing Gemini-2.5-Flash as the equivalence judge. In all scenarios, the evaluating model remains completely independent of both the task agent and the memory agent to prevent self-grading artifacts.

\subsection{Provider Implementation Parameters}
\label{appendix:provider_params}

All providers share a unified interface (\texttt{retrieve} and \texttt{store}) but differ in their internal representations and retrieval mechanisms. Table~\ref{tab:provider_params} summarizes these parameters. To prevent redundant loading, the 11 providers utilizing embeddings share a single 384-dimensional \texttt{all-MiniLM-L6-v2} encoder through a GPU-locked singleton.

Four providers require LLM calls during retrieval: Agent-KB for query refinement and guidance synthesis, LightWeight for memory selection and synthesis, Generative for per-candidate importance scoring (which falls back to cosine similarity if an LLM is unavailable), and Cheatsheet for curator synthesis. Conversely, Memp and EvolveR invoke LLMs exclusively during storage. Their retrieval paths rely entirely on cosine similarity search over embeddings without any generation step. Furthermore, all 13 providers necessitate LLM calls during storage to perform trajectory summarization, insight extraction, or skill generation.

\textbf{Summary.} For most providers, storage costs intentionally dominate retrieval costs because storage LLM calls are amortized across numerous subsequent retrievals. Additionally, the lightweight probe stage used for content-aware \textsc{Begin} routing completely bypasses the retrieval LLM for all 13 providers. By relying solely on embedding or keyword lookups, this stage ensures that probe latency remains bounded, regardless of a provider's full retrieval cost.

\begin{table}[htbp]
\centering
\caption{Implementation parameters for all 13 long-term memory providers. ``Emb.'': uses a dense embedding model on the retrieval path; ``LLM-R'': requires an LLM call on the full retrieval path; $k$: retrieval top-$k$, where $3{\to}1$ denotes three dense candidates reranked by an LLM to one, and $2{+}2$ denotes two tips plus two shortcuts. All 13 providers additionally invoke an LLM during storage. The content-aware \textsc{Begin} probe stage is embedding/keyword-only for all providers (no LLM).}
\label{tab:provider_params}
\resizebox{\textwidth}{!}{%
\begin{tabular}{lcccp{7cm}}
\toprule
\textbf{Provider} & \textbf{Emb.} & \textbf{$k$} & \textbf{LLM-R} & \textbf{Key Parameters} \\
\midrule
Voyager~\cite{wang2024voyager} & \ding{51} & 1 & \ding{55} & Pure semantic search; returns 6-sentence max trajectory summary with hierarchical distillation \\
ExpeL~\cite{zhao2024expel} & \ding{51} & 3 & \ding{55} & Hybrid search: TF-IDF 0.3, semantic 0.7; dual-path insight + trajectory retrieval \\
Generative~\cite{park2023generative} & \ding{51} & 3$\to$1 & \ding{51} & Semantic search returns 3 candidates; LLM importance scoring selects top-1; $<$150 words output \\
DILU~\cite{wen2024dilu} & \ding{51} & 1 & \ding{55} & Pure semantic search on query embeddings; ultra-concise step format ($<$150 words) \\
AWM~\cite{wang2025agent} & \ding{51} & 1 & \ding{55} & Pure embedding search; abstract workflow templates with entity generalization; $<$200 words \\
Mobile-E~\cite{wang2025mobile} & \ding{55} & 2+2 & \ding{55} & Jieba keyword tokenization; 2 tips + 2 shortcuts; structured action sequences \\
Cheatsheet~\cite{suzgun2026dynamic} & \ding{51} & 1 & \ding{51} & Global cheatsheet; LLM curator synthesizes from similar trajectories; $<$200 words per update \\
SkillWeaver~\cite{zheng2025skillweaver} & \ding{55} & 3 & \ding{55} & Keyword-based skill matching; safety validation blocks \texttt{exec/eval/import}; returns source code \\
Agent-KB~\cite{tang2025agent} & \ding{51} & 3 & \ding{51} & Hybrid search weights: TF-IDF 0.5, semantic 0.5; structured fields: planning, experience \\
Memp~\cite{fang2025memp} & \ding{51} & 1 & \ding{55} & Procedural scripts ($<$100 words) + concrete steps ($<$150 words); self-correction on failure \\
EvolveR~\cite{wu2025evolver} & \ding{51} & 1 & \ding{55} & Principle retrieval; merge threshold 0.8; prune threshold 0.3; max 1 pos + 1 neg example \\
LightWeight~\cite{zhang2025memevolve} & \ding{51} & 3 & \ding{51} & Dual-layer: max 30 strategic + 30 operational memories; short-term buffer size 20 \\
InsightGraph~(Ours) & \ding{51} & 5 & \ding{55} & Graph expansion (see Appendix~\ref{appendix:insight_graph}); 500-char output cap \\
\bottomrule
\end{tabular}%
}
\end{table}

\section{Method Details}
\label{appendix:method_details}

This section describes the end-to-end inference pipeline, the LLM-as-Judge reward design used for training data synthesis, the two memory providers we contribute (InsightGraph and the short-term memory provider), and the full decision prompt templates and output schemas.

\subsection{Inference Pipeline}
\label{appendix:inference_pipeline}

Algorithm~\ref{alg:inference} summarizes the end-to-end inference pipeline. For each task, the memory agent first probes all providers in parallel, routes to one for \textsc{Begin} retrieval, decides \textsc{Query}/\textsc{NoOp} for short-term memory at every step, and finally selects a subset of providers to store the completed trajectory.

\begin{algorithm}[htbp]
\caption{MemAgent Inference Pipeline}
\label{alg:inference}
\begin{algorithmic}[1]
\Require Task description $q$; memory agent $\mathcal{A}_M$; provider pool $\mathcal{P}$; task agent $\mathcal{A}_T$
\Ensure Task answer $\hat{y}_\tau$

\Statex \textbf{--- \textsc{Begin}: long-term memory routing ---}
\State $\{\mathrm{probe}_i\} \gets \textsc{ParallelProbe}(\mathcal{P}, q)$ \Comment{snippet previews from all providers}
\State $p^* \gets \mathcal{A}_M\!\big(q,\; \textsc{Shuffle}(\{\mathrm{probe}_i\})\big)$ \Comment{content-aware provider selection}
\State Inject $p^*\!.\textsc{Retrieve}(q)$ into $\mathcal{A}_T$'s system prompt

\Statex \textbf{--- \textsc{In}: short-term memory gating ---}
\State $\mathcal{M}_{\mathrm{stm}} \gets \emptyset$
\For{$t = 0, 1, \ldots, T$}
    \State $\mathcal{M}_{\mathrm{stm}} \gets \textsc{UpdateSTM}(\mathcal{M}_{\mathrm{stm}},\; a_{<t},\; o_{<t})$ \Comment{extract facts, URLs, searches}
    \State $d^{(t)} \gets \mathcal{A}_M\!\big(q,\; t,\; \mathcal{M}_{\mathrm{stm}},\; \{(a_i, o_i)\}_{i<t}^{\text{recent}}\big)$ \Comment{\textsc{Query} or \textsc{NoOp}}
    \If{$d^{(t)} = \textsc{Query}$ \textbf{and} $\mathcal{M}_{\mathrm{stm}} \neq \emptyset$}
        \State Inject $\textsc{Format}(\mathcal{M}_{\mathrm{stm}})$ into $\mathcal{A}_T$'s prompt
    \EndIf
    \State $(a_t, o_t) \gets \mathcal{A}_T.\textsc{Step}(t)$ \Comment{break if $a_t$ is terminal}
\EndFor
\State $\hat{y}_\tau \gets \mathcal{A}_T$'s final answer

\Statex \textbf{--- \textsc{End}: selective multi-store ---}
\State $\mathcal{S} \gets \mathcal{A}_M\!\big({y}_\tau,\; \textsc{Summarize}(\xi_\tau)\big)$ \Comment{select provider subset}
\For{each $p_j \in \mathcal{S}$} $p_j.\textsc{Store}(\xi_\tau)$ \EndFor
\end{algorithmic}
\end{algorithm}

\subsection{LLM-as-Judge Reward Design}
\label{appendix:judge_dimensions}

The per-decision reward model uses an LLM judge to evaluate each memory routing decision along phase-specific attribution dimensions. All dimensions are scored on a 0--5 integer scale and normalized to $[0,1]$ before aggregation. Dimensions marked with $\downarrow$ are \emph{inverted} (higher raw scores correspond to lower rewards). Table~\ref{tab:judge_dimensions} summarizes all dimensions and their weights.

\begin{table}[htbp]
\centering
\caption{LLM-as-Judge scoring dimensions and weights by decision phase. Dimensions marked $\downarrow$ are inverted before aggregation: $\hat{s}_i = 1 - s_i / s_{\max}$.}
\label{tab:judge_dimensions}
\resizebox{\textwidth}{!}{%
\begin{tabular}{llclp{7cm}}
\toprule
\textbf{Phase} & \textbf{Dimension} & \textbf{Weight} & \textbf{Dir.} & \textbf{Description} \\
\midrule
\multirow{5}{*}{\textsc{Begin}}
  & Utilization & 0.6 & $\uparrow$ & Did the task agent reference or use information from the retrieved memory in its subsequent actions? \\
  & Relevance   & 0.4 & $\uparrow$ & How relevant was the retrieved memory to the specific task? \\
\midrule
\multirow{4}{*}{\textsc{In} (\textsc{Query})}
  & Helpfulness & 0.4 & $\uparrow$ & Did the agent use information from the injected short-term memory? \\
  & Progress    & 0.4 & $\uparrow$ & Did the injection help advance the task? \\
  & Noise       & 0.2 & $\downarrow$ & Did the injection distract or confuse the agent? \\
\midrule
\multirow{6}{*}{\textsc{In} (\textsc{NoOp})}
  & Appropriateness      & 0.5 & $\uparrow$ & Was skipping retrieval a reasonable choice given the current state? \\
  & Progress             & 0.3 & $\uparrow$ & Did the task continue to make progress without memory retrieval? \\
  & Missed Opportunity   & 0.2 & $\downarrow$ & Did skipping retrieval cause the agent to miss important information? \\
\bottomrule
\end{tabular}%
}
\end{table}

\paragraph{Scoring rubrics.}
Each dimension uses a 6-level rubric (0--5). Table~\ref{tab:rubric_anchors} provides representative anchors across all phases. The general principle is consistent: a score of 0 indicates no effect (memory ignored, no progress, no missed information), 2--3 marks a borderline case, and 5 represents an unambiguous strong signal. For inverted dimensions (Noise, Missed Opportunity), the anchoring is reversed so that 0 corresponds to the best outcome.

\begin{table}[htbp]
\centering
\caption{Representative rubric anchors for selected dimensions. All dimensions follow a 0--5 integer scale. Dimensions marked $\downarrow$ are inverted before aggregation.}
\label{tab:rubric_anchors}
\resizebox{\textwidth}{!}{%
\begin{tabular}{llp{3.8cm}p{3.8cm}p{3.8cm}}
\toprule
\textbf{Phase} & \textbf{Dimension} & \textbf{0 (lowest)} & \textbf{2--3 (borderline)} & \textbf{5 (highest)} \\
\midrule
\textsc{Begin} & Utilization & Memory completely ignored & Some overlap, likely coincidental & Agent directly built upon specific memory content \\
\textsc{Begin} & Relevance & Completely irrelevant domain & Somewhat related, not specifically useful & Perfectly targeted, critical information \\
\midrule
\textsc{In} (Query) & Helpfulness & No reference to injected content & Minor overlap, uncertain attribution & Agent directly leveraged specific injected details \\
\textsc{In} (Query) & Noise $\downarrow$ & No distraction & Some off-topic exploration triggered & Agent clearly confused, pursued wrong direction \\
\midrule
\textsc{In} (NoOp) & Appropriateness & STM had highly relevant content that should have been retrieved & Borderline; STM has marginal content & STM empty or clearly irrelevant; NoOp optimal \\
\textsc{In} (NoOp) & Missed Opp. $\downarrow$ & Nothing valuable missed & Missed non-critical information & Missed critical information; retrieval would have been breakthrough \\
\bottomrule
\end{tabular}%
}
\end{table}

\paragraph{Aggregation.}
For each decision, the scalar reward is computed as
\begin{equation}
    r_d \;=\; \big\langle \mathbf{w}^{(\phi)},\, \hat{\mathbf{s}} \big\rangle
          \;=\; \sum_{i=1}^{K_\phi} w_i^{(\phi)}\, \hat{s}_i,
    \quad
    \hat{s}_i =
    \begin{cases}
      s_i / s_{\max}, & \text{dim.\ } i\uparrow \\[-2pt]
      1 - s_i / s_{\max}, & \text{dim.\ } i\downarrow
    \end{cases},
    \quad \mathbf{w}^{(\phi)} \in \Delta^{K_\phi - 1}
\end{equation}
where $s_{\max} = 5$, $\phi \in \{\textsc{Begin}, \textsc{In-Query}, \textsc{In-NoOp}\}$ indexes the decision phase, $K_\phi$ is the number of scoring dimensions for phase $\phi$, and $\mathbf{w}^{(\phi)}$ lies on the $(K_\phi{-}1)$-simplex so that $r_d \in [0,1]$. The judge outputs a JSON object with scores and a brief reasoning string. Parsing uses direct JSON decoding with a regex fallback for robustness.

\paragraph{Judge context windows.}
To control cost and focus the judge's attention, input fields are truncated: task description (500 char for \textsc{Begin}, 300 for \textsc{In}), retrieved memory content (800 char), subsequent actions (2000 char for \textsc{Begin}), and before/after actions (400 char each for \textsc{In}).

\subsection{InsightGraph: Adaptive Graph Memory}
\label{appendix:insight_graph}

InsightGraph is a graph-structured long-term memory provider designed to address two limitations of flat-list retrieval: the absence of relational structure between co-occurring strategies, and the inability to perform multi-hop associative recall. We describe its representation, retrieval mechanism, and maintenance protocol below.

\subsubsection{Graph Representation}

Memories are organized as a typed knowledge graph $\mathcal{G} = (\mathcal{V}, \mathcal{E})$. Each node $v \in \mathcal{V}$ stores a concise, actionable insight extracted from a successful trajectory (\eg, \emph{``For spreadsheet tasks, call inspect\_file\_as\_text rather than attempting web download''}), together with metadata linking it to its source task. Nodes are deduplicated using content-based signatures.

Edges capture two complementary relations. \emph{Same-task} edges ($w{=}1.0$) link insights co-occurring within one trajectory, encoding the intuition that strategies applied together to a single task are likely jointly relevant. \emph{Similar-concept} edges link semantically related insights \emph{across} trajectories: after a new node is inserted, its sentence embedding is compared against all existing nodes, and bidirectional edges are created for pairs exceeding a cosine similarity threshold $\tau_{\text{sim}} = 0.55$, with initial weight set to the similarity score. Each edge additionally maintains usage and success counters $(n_e, k_e)$ for online weight adaptation.

\subsubsection{Retrieval}

Retrieval operates at the \textsc{Begin} phase and combines sparse lexical matching with dense semantic search and graph expansion.

\paragraph{Seed selection.}
Given a task query $q$, a hybrid score blends TF-IDF and embedding similarity:
\begin{equation}
    s(v, q) = 0.2 \cdot \cos(\text{TF-IDF}_q, \text{TF-IDF}_v) + 0.8 \cdot \cos(\mathbf{e}_q, \mathbf{e}_v)
\end{equation}
The dense term dominates because task queries are typically short and benefit from semantic generalization, while the sparse term handles exact entity matches (tool names, domain keywords). The top-5 nodes form the seed set $\mathcal{S}$.

\paragraph{Graph expansion.}
Seed scores are propagated to neighbors to surface associatively related insights that would be missed by direct retrieval:
\begin{equation}
    s(u, q) \leftarrow \max\!\big(s(u, q),\; s(v, q) \cdot w_{\mathrm{eff}}(e_{v \to u}) \cdot \gamma\big), \quad v \in \mathcal{S}
\end{equation}
with decay $\gamma = 0.7$ and a maximum of 3 neighbors per seed. Single-hop expansion is sufficient in practice: the graph remains relatively sparse (${\sim}$50--200 nodes after typical training), and deeper expansion surfaces tangentially related content.

\paragraph{Composition.}
Nodes scoring below $\tau_{\min} = 0.22$ are discarded. The top-3 survivors are concatenated into a guidance text. No LLM call is made on the retrieval path, keeping latency negligible.

\subsubsection{Adaptive Edge Weights}

Edge weights reflect empirical co-utility, defined as the frequency with which traversing an edge leads to a successful task outcome. This approach ensures that weights do not rely solely on the initial semantic similarity estimate. Specifically, we blend a semantic prior $w_0$ with a Bayesian quality term as follows:
\begin{equation}
    w_{\mathrm{eff}}(e) = \alpha_e \cdot w_0(e) + (1 - \alpha_e) \cdot \frac{k_e + 1}{n_e + 2}, \qquad \alpha_e = \frac{1}{1 + n_e / \kappa}
\end{equation}
The quality term employs Laplace smoothing with a $\text{Beta}(1,1)$ prior. It starts at 0.5 for unused edges and converges to the empirical success rate $k_e / n_e$ as evidence accumulates. A half-life parameter, $\kappa = 5$, controls the blending coefficient $\alpha_e$. This coefficient transitions from being dominated by the prior during a cold start ($\alpha_e \approx 1$) to being dominated by experience after approximately 15 uses ($\alpha_e \approx 0.25$). This dual-phase behavior prevents the premature pruning of promising but underexplored edges while reducing the weight of thoroughly tested, poorly performing edges. The system updates these weights after each task and clamps them to the range $[0.1, 1.0]$. To prevent overfitting from a single observation, updates only occur when $n_e \geq 2$.

\subsubsection{Storage and Consolidation}

Upon successful task completion, an LLM extracts 2--4 concise insights ($\leq$30 words each, imperative form) from the trajectory. Each insight becomes a new node; same-task and similar-concept edges are created as described above.

To bound graph growth, periodic consolidation runs every 50 tasks: near-duplicate nodes (cosine similarity $\geq 0.85$) are merged, and poorly performing similar-concept edges ($n_e \geq 5$, success rate ${<}\,20\%$) are pruned. This keeps the graph compact while retaining empirically validated associations.

\subsection{Short-Term Memory Provider}
\label{appendix:short_term_memory}

Prior work on memory-augmented agents focuses primarily on long-term, cross-task memory. We contribute a complementary \emph{short-term} provider that aggregates within-task working state. The task agent already has access to its full conversation history, but cannot efficiently derive cross-step summaries (\eg, \emph{which URLs have been visited?}, \emph{what facts have been verified so far?}) by scanning raw tool-call logs. The short-term provider is designed to fill this gap under three constraints: extraction is rule-based, so the retrieval path has deterministic behavior and negligible latency and makes no LLM calls; the output does not duplicate raw observations already present in the conversation history; and the provider adopts a conservative display policy that is intended to reduce low-information injections when insufficient state has accumulated.

\subsubsection{Extraction and State Management}

At each step, a rule-based extractor parses the structured tool-call record and updates three bounded data structures:

\begin{enumerate}[leftmargin=*]
\item \textbf{Verified facts} (capacity: 15 active). We employ five regular expression patterns to extract common fact formats. These include key-value pairs with numeric values (\eg, ``Population: 21.54 million''), explicit answer statements, founding dates, geographic coordinates, and ``X is Y'' assertions. This extraction process yields a maximum of three facts per step. To prevent redundancy, a word-level Jaccard similarity threshold ($>0.6$) filters out near-duplicates. Finally, if the memory capacity is exceeded, the oldest facts are evicted.

\item \textbf{URL history} (capacity: 30). Each visited or discovered URL is recorded alongside its associated tool, step index, and outcome (success or error). The eviction policy prioritizes unvisited URLs (FIFO), followed by the oldest entries overall.

\item \textbf{Search history} (capacity: 20). Each search query is logged with its step index and result count, employing a FIFO eviction policy.
\end{enumerate}

Error detection runs in parallel. Fourteen compiled regex patterns classify tool failures into specific categories (\eg, access denied, not found, timeout, rate limit, and no results). To avoid false positives from normal page content, this process scans only the first 500 characters of each observation.

\subsubsection{Output Formatting}

The formatter renders working memory into up to three prioritized sections (Table~\ref{tab:stm_sections}): Collected Data ($\geq 2$ facts to display), Pages Visited ($\geq 3$ URLs), and Recent Searches ($\geq 4$ queries). Under the conservative display policy, when all sections fall below their minimum thresholds the provider returns an empty response rather than rendering sparse content. When the character budget is exceeded, truncation occurs at the last complete section boundary so that no section is partially rendered.

\begin{table}[htbp]
\centering
\caption{Short-term memory output sections, display thresholds, and capacity limits.}
\label{tab:stm_sections}
\begin{tabular}{lccc}
\toprule
\textbf{Section} & \textbf{Min to Show} & \textbf{Max Items} & \textbf{Priority} \\
\midrule
Collected Data   & 2 facts   & 8  & Highest \\
Pages Visited    & 3 URLs    & 15 & Medium \\
Recent Searches  & 4 queries & 8  & Lowest \\
\bottomrule
\end{tabular}
\end{table}

\subsubsection{Worked Example}

We illustrate the formatter's behavior on a small input trace.

\textbf{Input (3 steps).}
\begin{quote}\small
\texttt{Step 1} \textemdash{} \texttt{web\_search("country X population 2024")} $\to$ observation containing \texttt{"Population: 21.54 million (source: wikipedia.org)"} \\
\texttt{Step 2} \textemdash{} \texttt{crawl\_page("https://en.wikipedia.org/wiki/X")} $\to$ observation containing \texttt{"GDP growth: 3.2\%"} \\
\texttt{Step 3} \textemdash{} \texttt{web\_search("country X GDP 2024")} $\to$ 3 results (no new extractable fact)
\end{quote}

\textbf{Output (rendered).}
\begin{quote}\small
\texttt{>> COLLECTED DATA:} \\
\texttt{~~- Population: 21.54 million (source: wikipedia.org)} \\
\texttt{~~- GDP growth: 3.2\%} \\
\texttt{:: PAGES VISITED: (suppressed; 1 URL < 3 threshold)} \\
\texttt{?? RECENT SEARCHES: (suppressed; 2 queries < 4 threshold)}
\end{quote}

Here Collected Data meets its 2-fact threshold and is displayed. The Pages Visited and Recent Searches sections each fall below threshold and are suppressed, so the final output contains only the verified facts. Had only one fact been extracted with no URLs and one search, all three sections would fall below threshold and the provider would return an empty response, deferring entirely to the agent's conversation history rather than injecting sparse content.

\subsection{Decision Prompt Format and Output Schema}
\label{appendix:prompt_templates}

All three decision prompts share a common structure comprising a role instruction, a provider description guide, phase-specific context fields, decision rules, and a JSON-only output constraint. These prompts remain identical during both training and inference. The three phases differ in their input fields, output schemas, and allowed action sets.

During the \textsc{Begin} phase, the prompt includes the task description, the retrieval query, a randomly shuffled provider guide, and shuffled probe results from all 13 providers with their scores removed. Here, the action is strictly constrained to \texttt{query}. The agent must output \texttt{\{"action": "query", "provider": <one of 13>\}} and is explicitly instructed not to answer the task itself.
For the \textsc{In} phase, the prompt contains the task question, the current step index, a short-term memory summary, and recent tool-call observations. The agent makes a binary decision formatted as \texttt{\{"action": "query"|"no\_op", "provider": "short\_term\_memory"|null\}}. It prefers \texttt{query} when the short-term memory holds useful, otherwise inaccessible content, and defaults to \texttt{no\_op} otherwise.

In the \textsc{End} phase, the prompt provides the task outcome, a trajectory summary, and the provider guide annotated with outcome gates. The agent outputs \texttt{\{"selected\_providers": [<subset of 13>]\}}, which typically includes three to six providers, though an empty list is permitted. Finally, Table~\ref{tab:provider_descriptions} details the provider guide. This guide serves as the definitive reference for each provider's storage and retrieval behavior during the \textsc{Begin} and \textsc{End} decisions.

\begin{table}[htbp]
\centering
\caption{Provider descriptions injected into the memory agent prompts. The \textsc{Begin} prompts incorporate both storage and retrieval descriptions, whereas the \textsc{End} prompts include only the storage description accompanied by an outcome gate annotation.}
\label{tab:provider_descriptions}
\resizebox{\textwidth}{!}{%
\begin{tabular}{lp{6.2cm}p{6.2cm}p{1.8cm}}
\toprule
\textbf{Provider} & \textbf{Storage Description} & \textbf{Retrieval Description} & \textbf{Gate} \\
\midrule
LightWeight & Extracts strategic (planning/decisions) and operational (tools/techniques) memories after task success & Provides long-term strategic/operational experience & Success \\
Agent-KB & Summarizes trajectories into structured fields: Agent Planning, Search Agent Planning, Agent Experience & Student Guidance (reference plans) and Teacher Guidance (lessons learned) & Success \\
SkillWeaver & Analyzes trajectories to identify core algorithms, generates generalized Python functions & Matches relevant skills, dynamically wraps them as tools & Success \\
ExpeL & Extracts Insights (Tips or Warnings) from all outcomes; stores refined trajectories on success & Hybrid retrieval of Insights and successful trajectories & All \\
Voyager & Generates trajectory summaries from all outcomes, stored as JSON with embeddings & Semantic search for most similar historical tasks & All \\
DILU & Stores query embeddings and trajectory summaries from both successes and failures & Semantic search on query similarity & All \\
Generative & Stores task queries and trajectory summaries & Vector retrieval + LLM importance scoring & All \\
Memp & Induces process scripts on success; triggers memory adjustment on failure & Retrieves the most similar process scripts & All \\
Cheatsheet & Maintains globally updated cheatsheet of best practices and rules & Uses LLM as Curator to synthesize/update the cheatsheet & All \\
Mobile-E & Stores Tips (text) and Shortcuts (structured sequences) for device automation & Keyword-based retrieval of Tips and Shortcuts & Tips: All; Shortcuts: Success \\
AWM & Induces abstract, general workflow templates from specific trajectories & Vector search for the most similar workflows & Success \\
EvolveR & Extracts principles with semantic merge/prune; periodically prunes low-value ones & Retrieves relevant principles with positive/negative examples & All \\
InsightGraph & Extracts concise insights as graph nodes with semantic edges; periodic consolidation & Hybrid TF-IDF + semantic search $\to$ graph expansion & Success \\
\bottomrule
\end{tabular}%
}
\end{table}

\subsubsection{Full Decision Prompts}

We provide the verbatim prompt templates used at inference. Placeholders in braces (\eg, \texttt{\{task\_question\}}) are substituted at runtime.

\begin{lstlisting}[basicstyle=\ttfamily\scriptsize,breaklines=true,frame=single,caption={\textsc{Begin} phase decision prompt.}]
You are a Memory Decision Agent. Your ONLY task is to
output a JSON decision object.

[Available Memory Providers]
### lightweight_memory
- Storage: Extracts strategic and operational memories...
- Retrieval: Provides long-term strategic/operational experience.
### expel
- Storage: Extracts Insights (Tips/Warnings) from all outcomes...
- Retrieval: Hybrid retrieval of Insights and trajectories...
... (all 13 providers, randomly shuffled)

[Task Context]
- Task ID: {task_id}
- Phase: BEGIN
- Step: 0

[Task Description]
{task_question}

[Memory Request]
- Type: retrieval
- Query: {query}

[Memory Probe Results -- Actual Content Preview]
(Each provider was probed with your task query.)
### memp (42 items)
  Use web_search to find the target page, then crawl...
### expel (18 items)
  Insight: For spreadsheet tasks, use inspect_file...
... (all 13 providers, randomly shuffled, no scores)

[Decision Rules for RETRIEVAL]
- Phase is BEGIN: you MUST return "query" to retrieve
  relevant memories.

[CRITICAL INSTRUCTION]
You MUST respond with ONLY a valid JSON object.
Do NOT answer the task question.
You MUST choose a provider from the available list.

Output format:
{"action": "query", "provider": "<provider_name>"}

Your JSON response:
\end{lstlisting}

\begin{lstlisting}[basicstyle=\ttfamily\scriptsize,breaklines=true,frame=single,caption={\textsc{In} phase decision prompt.}]
[Task Context]
Task ID: {task_id}
Original Question: {task_question}
Current Phase: IN (Step {step_index})

[Short-Term Memory (Current Task)]
>> COLLECTED DATA:
  - Population: 21.54 million (source: wikipedia.org)
  - GDP growth: 3.2% (source: worldbank.org)
:: PAGES VISITED:
  - [OK] https://en.wikipedia.org/wiki/... (crawl_page)
  - [OK] https://data.worldbank.org/... (crawl_page)
?? RECENT SEARCHES:
  - "country X population 2024" (3 results)

[Recent Actions/Observations]
  1. [tool_call] web_search("country X GDP growth")
  2. [observation] Found 3 results: ...

[Available Memory]
  - short_term_memory: Structured working memory tracking
    cross-step verified facts, visited URLs, and search
    history.

[Decision Rules]
- Choose QUERY when:
  * Short-term memory has useful content.
  * The task agent would benefit from recalling earlier steps.
- Choose NO_OP when:
  * Short-term memory is empty.
  * The agent already has all necessary context visible.

[Decision Required]
1. QUERY: Retrieve structured state from short-term memory.
2. NO_OP: Skip memory injection for this step.

{"action": "query" | "no_op", "provider": "short_term_memory" | null}
\end{lstlisting}

\begin{lstlisting}[basicstyle=\ttfamily\scriptsize,breaklines=true,frame=single,caption={\textsc{End} phase multi-store prompt.}]
You are a Memory Storage Decision Agent. Your task is to
decide which memory providers should store this trajectory.

[Task Result]: SUCCESS
[Task Query]: {task_question}

[Trajectory Summary]:
{trajectory_summary}

[Available Memory Providers]:
  - expel: extracts Insights from both successes/failures.
    [All outcomes]
  - lightweight_memory: stores strategic and operational
    experience. [Success-only]
  - skillweaver: converts trajectories into reusable Python
    functions. [Success-only]
  ... (all 13 providers, randomly shuffled)

[Decision Guidelines]:
1. You can select MULTIPLE providers (recommended 3-6)
2. If content is not valuable, return an empty list

[Output Format]:
{"selected_providers": ["provider1", "provider2", ...]}

Your JSON response:
\end{lstlisting}

\subsubsection{LLM-as-Judge Prompts}

The LLM-as-Judge prompts assess the quality of each routing decision for training data synthesis. Here, we present the \textsc{Begin} prompt. The \textsc{In} (\textsc{Query}) and \textsc{In} (\textsc{NoOp}) prompts follow an analogous structure, incorporating their respective evaluation dimensions detailed in Table~\ref{tab:judge_dimensions}.

\begin{lstlisting}[basicstyle=\ttfamily\scriptsize,breaklines=true,frame=single,caption={LLM-as-Judge prompt for \textsc{Begin} phase scoring.}]
You are a memory quality evaluator for an AI agent system.

A Memory Agent retrieved historical memory to help a Task
Agent solve a task. Evaluate whether this retrieval was useful.

## Task
{task_context}

## Retrieved Memory Content
{memory_content}

## Task Agent's Subsequent Actions
{subsequent_actions}

## Evaluation Criteria
Score each dimension from 0-5:

1. **Utilization** (0-5): Did the Task Agent reference or
   use information from the retrieved memory?
   - 0: Memory completely ignored
   - 2: Some overlap, likely coincidental
   - 4: Agent clearly referenced memory content
   - 5: Agent directly quoted or built upon memory

2. **Relevance** (0-5): How relevant was the retrieved
   memory to this specific task?
   - 0: Completely irrelevant (different domain)
   - 2: Somewhat related but not specifically useful
   - 4: Highly relevant with directly applicable strategies
   - 5: Perfectly targeted, critical information

## Output Format
{"utilization": <0-5>, "relevance": <0-5>,
 "reasoning": "<brief explanation>"}
\end{lstlisting}

\subsubsection{Decision Output Format}

The structured output of the memory agent differs across phases. During the \textsc{Begin} and \textsc{In} phases, the agent generates a single-provider JSON object:
\begin{verbatim}
{"action": "query", "provider": "expel"}
{"action": "no_op", "provider": null}
\end{verbatim}
For the \textsc{End} phase, the output is a JSON object containing a list of multiple selected providers:
\begin{verbatim}
{"selected_providers": ["expel", "voyager", "dilu",
                        "lightweight_memory", "agent_kb"]}
\end{verbatim}

\section{Further Analyses}
\label{appendix:additional_analyses}

This section provides complementary analyses: we detail the training-free PromptRoute baseline and its Decision Knowledge Base, dissect MemAgent's routing behavior across the three decision phases, and quantify the prompt-level injection overhead relative to individual memory providers. We additionally examine how router capacity affects performance.

\subsection{Router Capacity}
\label{appendix:router_capacity}

We scale the memory-agent backbone from Qwen3.5-4B to Qwen3.5-9B while retaining the same training and evaluation protocol. \cref{tab:router_capacity} reports both untrained and fine-tuned variants on GAIA.

\begin{table}[htbp]
\centering
\caption{Router-capacity analysis on GAIA. Accuracy (\%).}
\label{tab:router_capacity}
\small
\setlength{\tabcolsep}{4pt}
\begin{tabular}{lcccc}
\toprule
\textbf{Method} & \textbf{Overall} & \textbf{L1} & \textbf{L2} & \textbf{L3} \\
\midrule
Qwen3.5-4B w/o Training & 63.6 & 71.7 & 67.4 & 34.6 \\
MemAgent (Qwen3.5-4B)   & 69.7 & \textbf{83.0} & 67.4 & 50.0 \\
Qwen3.5-9B w/o Training & 63.6 & 75.5 & 61.6 & 46.2 \\
MemAgent (Qwen3.5-9B)   & \textbf{70.9} & 81.1 & \textbf{69.8} & \textbf{53.8} \\
\bottomrule
\end{tabular}
\end{table}

Increasing router capacity yields a limited but consistent overall improvement after training: the 9B variant raises overall accuracy from 69.7\% to 70.9\% and Level-3 accuracy from 50.0\% to 53.8\%. The untrained 4B and 9B variants obtain the same overall accuracy, indicating that the gain arises from the interaction between capacity and routing supervision rather than model size alone.

\subsection{PromptRoute: Decision Knowledge Base}
\label{appendix:promptroute_dkb}

PromptRoute serves as a training-free baseline that replaces the SFT-trained memory agent with a frozen API LLM (GPT-5-mini). It augments the routing prompt with a Decision Knowledge Base (DKB), which functions as a structured JSON knowledge store. For each of the 13 providers, the DKB records empirical success rates, primary task-type affinities, and representative examples of successful and failed routing. Furthermore, it tailors information to specific phases. During the \textsc{Begin} phase, the DKB encodes routing rules that map task types to providers, subject to a minimum evidence threshold ($\geq 3$ occurrences and a win rate of $\geq 40\%$). For the \textsc{In} phase, it supplies simple heuristics (\eg, querying when the short-term memory contains content and skipping when it is empty). Finally, for the \textsc{End} phase, it outlines provider storage compatibility (\eg, success-only versus all outcomes). At inference time, the relevant subset of the DKB is serialized and integrated into the routing prompt alongside the provider guide and probe results.

The DKB is initialized through a one-time bootstrapping process using scored trajectories generated during forced exploration. A keyword-based classifier assigns each task to one of eight categories (\eg, image analysis, web search, and calculation), after which provider-level statistics and routing rules are aggregated. This entire process requires no GPU training, as it relies solely on LLM API calls for the initial exploration. Ultimately, PromptRoute achieves a 65.5\% accuracy on GAIA (+6.7\% over the no-memory baseline) with zero training cost, reaching 94\% of the performance of the trained model (69.7\%). The remaining performance gap suggests that applying SFT on per-decision rewards captures finer-grained routing distinctions than prompt-based knowledge retrieval. Nevertheless, PromptRoute remains a promising alternative for future research.

\subsection{Routing Behavior}
\label{appendix:routing_behavior}

We examine MemAgent's routing decisions across the three decision phases (\textsc{Begin}, \textsc{In}, \textsc{End}) and the three benchmarks.

\subsubsection{Phase-Specific Routing Distribution}

\paragraph{\textsc{Begin} phase.}
The model routes across all 13 providers rather than concentrating on a single dominant option. The top four providers are Memp 22.6\%, DILU 19.0\%, Mobile-E 16.7\%, and ExpeL 16.7\%, with the remaining nine sharing the balance. All 13 providers are selected at least once, and no JSON parse failures occur on the evaluation set.

\paragraph{\textsc{In} phase.}
Across the 165 GAIA evaluation tasks (comprising 1{,}102 \textsc{In}-phase decisions in total), the memory agent selects \textsc{Query} on approximately 30.4\% of the steps and \textsc{NoOp} on 69.6\%. The high \textsc{NoOp} rate is largely driven by early steps. For instance, Step~0 defaults to \textsc{NoOp} because the short-term memory is empty at the start of a task. As short-term memory accumulates, the \textsc{Query} rate rises to roughly 55\% to 71\% by Steps~5 and 6 before eventually tapering off. The memory agent consistently selects \textsc{NoOp} whenever the short-term memory is empty or its contents are already reflected in the recent conversation.

\paragraph{\textsc{End} phase.}
MemAgent stores trajectories in three to eight providers rather than all 13. This selectivity is crucial, as indiscriminate storage pollutes retrieval pools with low-relevance memories. Upon a successful task outcome, storage includes both all-outcomes providers (\eg, ExpeL, Voyager, and DILU) and success-only providers (\eg, LightWeight, Agent-KB, and InsightGraph). Conversely, upon failure, the memory agent selects only all-outcomes providers.

\subsubsection{Probe Scores}
We observe that including numerical probe scores in the \textsc{Begin} prompt induces a highly skewed selection distribution. When these scores are present, 51.8\% of selections are concentrated on just two providers, even though their score magnitudes are not directly comparable. Conversely, removing the scores and randomly shuffling the probe list yields a more uniform distribution with no loss in downstream accuracy. This behavior suggests that the agent inherently treats numerical scores as a direct ranking signal rather than a content preview; omitting them forces the agent to rely entirely on the retrieved content itself.

\subsubsection{Provider Performance Specialization}

Provider coverage and specialization across benchmarks reveal complementary strengths that motivate the routing problem.

\paragraph{Per-benchmark top performers.}

\begin{itemize}[leftmargin=*]
    \item \textbf{GAIA}: AWM (+6.1pp), Voyager (+5.5pp), SkillWeaver (+4.8pp), Mobile-E (+4.8pp), ExpeL (+3.0pp)
    \item \textbf{WebWalkerQA}: ExpeL (+7.6pp), Mobile-E (+4.7pp), LightWeight (+4.1pp), InsightGraph (+2.9pp), Generative (+2.9pp)
    \item \textbf{xBench DeepSearch}: SkillWeaver (+8.0pp), DILU (+6.0pp), InsightGraph (+6.0pp), ExpeL (+5.0pp), LightWeight (+4.0pp)
\end{itemize}

No single provider is optimal across all three benchmarks. AWM dominates on multi-step GAIA reasoning; ExpeL leads on web navigation (WebWalkerQA); SkillWeaver specializes in code-heavy search tasks (xBench-DS). This heterogeneity motivates the routing problem: a fixed provider choice will underperform a learned provider-per-task policy on at least one benchmark.

\paragraph{Task coverage analysis.}

Table~\ref{tab:coverage} illustrates how many of the 14 methods (13 memory providers and the no-memory baseline) individually solve each task. The three benchmarks exhibit distinct coverage profiles. GAIA is the most heterogeneous, featuring 12.1\% routing-critical tasks (solved by only one or two providers) and 35.2\% trivial tasks (solved by all 14). WebWalkerQA is skewed toward trivial tasks (51.2\%), with only 6.5\% classified as routing-critical. Conversely, xBench-DS contains the fewest routing-critical tasks (8.0\%) and is dominated by a broad band of tasks solved by 7 to 13 providers (44.0\%), indicating a moderate but widely covered difficulty level. Routing-critical tasks, where only a handful of providers succeed, represent the primary scenarios in which a memory agent can outperform any fixed choice.

\begin{table}[htbp]
\centering
\caption{Task coverage distribution: the number of methods (13 memory providers + no-memory baseline) that solve each task.}
\label{tab:coverage}
\begin{tabular}{lccc}
\toprule
\textbf{Correct} & \textbf{GAIA} & \textbf{WebWalkerQA} & \textbf{xBench-DS} \\
\midrule
0 (unsolvable)          & 12.7\% & 10.6\% & 6.0\% \\
1--2 (routing-critical) & 12.1\% & 6.5\%  & 8.0\% \\
3--6                    & 11.5\% & 8.2\%  & 7.0\% \\
7--13                   & 28.5\% & 23.5\% & 44.0\% \\
14 (all correct)        & 35.2\% & 51.2\% & 35.0\% \\
\bottomrule
\end{tabular}
\end{table}

\subsubsection{Failure Analysis}

\begin{figure*}[t]
\centering
\begin{subfigure}[t]{0.41\textwidth}
    \centering
    \includegraphics[width=\linewidth]{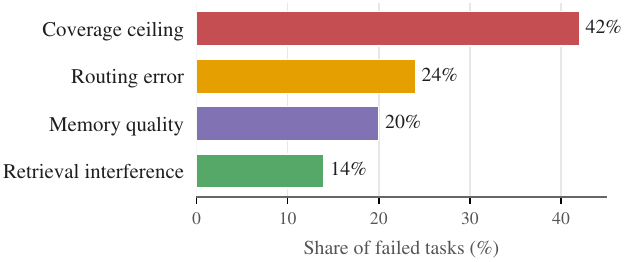}
    \vspace{-1em}
    \caption{GAIA failure root causes.}
    \label{fig:failure_cats}
\end{subfigure}
\hfill
\begin{subfigure}[t]{0.57\textwidth}
    \centering
    \includegraphics[width=\linewidth]{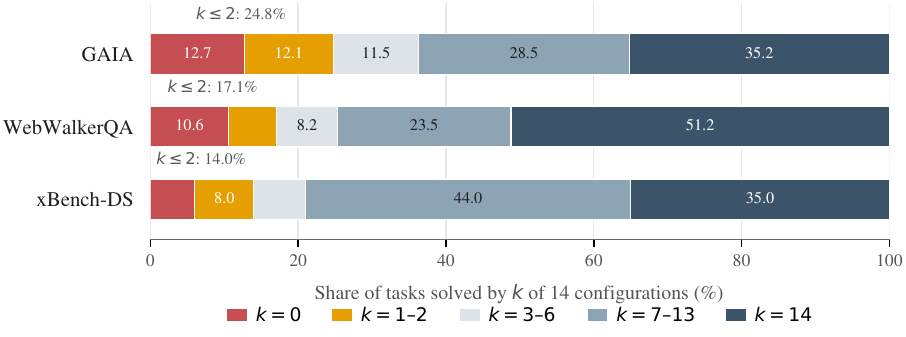}
    \vspace{-1em}
    \caption{Task solvability and number of successful methods $k$.}
    \label{fig:failure_coverage}
\end{subfigure}
\caption{\textbf{Failure analysis and task solvability across benchmarks.} Left: among failed tasks, coverage ceiling is the dominant failure source. Right: GAIA is more routing-critical than WebWalkerQA or xBench-DS, with more tasks solved by at most two of the 14 configurations.}
\label{fig:failure}
\end{figure*}

\cref{fig:failure}\subref{fig:failure_cats} categorizes MemAgent's failure modes on GAIA tasks. The dominant cause is a \emph{coverage ceiling} (42\%), representing tasks that neither the 13 providers nor the no-memory baseline can solve due to missing tool support or an inability to answer. The remaining failures fall into three categories. First, \emph{routing errors} (24\%) occur when MemAgent selects an incorrect option despite one or two providers being capable of success. Second, \emph{memory quality} issues (20\%) arise when the chosen provider is appropriate but yields suboptimal retrieved content. Third, \emph{retrieval interference} (14\%) happens when injected memory degrades performance on tasks the baseline normally solves. This mirrors the force-all-store ablation in \cref{tab:arch_ablation}, demonstrating that indiscriminate storage harms performance.

\cref{fig:failure}\subref{fig:failure_coverage} contextualizes these failures across benchmarks by binning tasks according to the number of successful providers ($k$). The ceiling-and-critical region ($k{\leq}2$) is largest on GAIA ($24.8\%$), followed by WebWalkerQA ($17.1\%$) and xBench-DS ($14.0\%$). On xBench-DS, $79\%$ of tasks are solved by at least seven providers, meaning routing primarily redistributes performance across a broadly solvable pool. In contrast, routing on GAIA more frequently operates near the edge of feasibility.

\subsection{Injection Overhead}
\label{appendix:injection_overhead}

To quantify the context-length cost of memory augmentation, Table~\ref{tab:injection_comparison} reports the number of tokens injected into the task-agent context during the \textsc{Begin} phase, together with the resulting average trajectory length on the 165 GAIA tasks. This measurement isolates the prompt-level overhead incurred at task initialization, and excludes both the router's decoding cost and any method-specific insertions that occur in later phases.

\begin{table}[htbp]
\centering
\caption{Added context size in the \textsc{Begin} phase and average trajectory length on GAIA. SkillWeaver is reported as ``tools only'' because it exposes retrieved skills as callable tools rather than injecting plain text, so a token count is not directly comparable.}
\label{tab:injection_comparison}
\begin{tabular}{lccc}
\toprule
\textbf{Method} & \textbf{Tokens injected} & \textbf{Avg.\ steps} & \textbf{$\Delta$ Steps} \\
\midrule
No Memory          & N/A          & 9.1  & N/A    \\
\midrule
InsightGraph       & 57           & 8.6  & $-$0.5 \\
LightWeight        & 154          & 8.3  & $-$0.8 \\
Voyager & 178         & 8.3  & $-$0.8 \\
DILU               & 184          & 8.3  & $-$0.8 \\
Generative         & 195          & 8.5  & $-$0.6 \\
Mobile-E           & 204          & 8.4  & $-$0.7 \\
Memp               & 286          & 8.6  & $-$0.5 \\
Cheatsheet         & 310          & 8.8  & $-$0.3 \\
AWM                & 341          & 8.7  & $-$0.4 \\
EvolveR            & 369          & 8.7  & $-$0.4 \\
Agent-KB           & 419          & 8.9  & $-$0.2 \\
ExpeL              & 971          & 8.4  & $-$0.7 \\
SkillWeaver        & tools only   & 8.3  & $-$0.8 \\
\midrule
\textbf{MemAgent}  & \textbf{369} & \textbf{8.0} & $\mathbf{-1.1}$ \\
\bottomrule
\end{tabular}
\end{table}

All memory methods yield shorter average trajectories than the no-memory baseline, confirming that retrieved context helps the task agent reach a solution in fewer steps. However, this relationship is non-monotonic: incorporating more tokens does not reliably translate into fewer steps, suggesting that performance gains are driven by the \emph{relevance} of the retrieved content rather than its sheer volume. MemAgent optimizes this trade-off, achieving the largest reduction in trajectory length ($-$1.1 steps, a 12\% relative decrease from the baseline) while adding only about 369 tokens per task. This count is well below Agent-KB (419) and less than half of ExpeL (971). These results demonstrate that learned routing enhances context efficiency by selecting \emph{useful} memories, rather than simply inflating the prompt size.

\section{Why Do Memory Providers Exhibit Complementary Behavior?}
\label{appendix:provider_failure_analysis}

No single memory provider dominates across the three evaluated benchmarks (\cref{tab:main_results}). This section offers a qualitative lens on that complementarity: each provider exposes a distinct representation and retrieval contract, which makes different kinds of past experience available for transfer. We (i) organize providers by their indexable units and characteristic failure modes and (ii) illustrate these differences through concrete cases drawn from the populated memory stores.

\subsection{Three Representational Families and Their Failure Modes}
\label{appendix:systematic_failures}

We group providers by their stored indexable units. A ``memory'' takes various forms: a trajectory summary in Voyager, a 30-word insight in InsightGraph, a callable function in SkillWeaver, or a running cheatsheet in Dynamic Cheatsheet. The inherent incomparability of these units necessitates a content-aware router. As summarized in \cref{tab:provider_families}, the following provider families each optimize for a different notion of cross-task reuse.

\begin{table}[htbp]
\centering
\small
\caption{\textbf{Three representational families.} Each family optimizes a different notion of cross-task reuse and, consequently, fails in different regimes.}
\label{tab:provider_families}
\begin{tabular}{p{2.8cm}p{4.2cm}p{5cm}}
\toprule
\textbf{Family} & \textbf{Members} & \textbf{Indexable unit} \\
\midrule
Similarity-retrieval & Voyager, DILU, Generative, Cheatsheet & Trajectory or query embedding \\
Structural-abstraction & AWM, SkillWeaver, Memp & Workflow template, executable function, or procedural script \\
Insight-accumulation & ExpeL, Agent-KB, InsightGraph, Mobile-E, EvolveR, LightWeight & Atomic, composable insight fragments \\
\bottomrule
\end{tabular}
\end{table}

\paragraph{Similarity-retrieval.}
These providers index trajectories by semantic similarity of either the task query or the trajectory summary. They succeed when the new task's surface form closely matches a stored task (\eg, repetitive ``watch YouTube link $\to$ transcribe $\to$ extract fact'' patterns). They fail in two regimes. \emph{(i) Surface paraphrase masks logical change}: two tasks can read almost identically yet impose different constraints (\eg, ``albums \emph{before} 2000'' vs.\ ``albums \emph{between} 2000--2009''); nearest-neighbour retrieval returns the closest surface match even when the underlying constraint is wrong. \emph{(ii) Domain shift with preserved structure}: when the new task's domain is unseen in storage, similarity retrieval cannot surface a relevant trajectory even if the reasoning skeleton matches a stored one. Dynamic Cheatsheet suffers an additional \emph{monotonicity bias}: it rewrites a single global cheatsheet each retrieval, so its content is dominated by whichever task type appeared most recently and early-domain information is implicitly forgotten.

\paragraph{Structural-abstraction.}
These providers deliberately discard task-specific content and retain only the shape of a solution: AWM parameterizes workflows with variable names (\texttt{target\_subject\_page}, \texttt{revision\_year}); SkillWeaver compiles generalized Python functions; Memp induces a ``mental-model'' script stripped of specific entities. They excel when two tasks share abstract computational logic despite unrelated surfaces. Their failure modes are complementary. \emph{Over-generalization} -- Memp's entity-stripped scripts lose the minutiae some tasks depend on. \emph{Under-coverage} -- SkillWeaver retains very few skills, so if the incoming task does not match an existing skill signature, retrieval returns nothing. \emph{Missing warnings} -- AWM ingests only successful trajectories; on tasks where the agent's natural tendency is a known pitfall, AWM has no mechanism to surface a caution.

\paragraph{Insight-accumulation.}
These providers decompose each trajectory into many small, reusable units; each unit is retrievable independently and composes with others. This is strongest on tasks whose success depends on \emph{assembling} several atomic best-practices (\eg, site-restricted search + cross-check against reference list + cite exact section). The failure modes are driven by retrieval geometry. \emph{Lexical flooding}: with thousands of stored insights, many share common nouns (``Wikipedia'', ``search''), so top-$k$ degenerates to generic advice and task-specific lessons drop below the cutoff. \emph{Keyword brittleness}: Mobile-E uses pure keyword matching, so queries that do not share surface vocabulary with stored tips retrieve nothing. \emph{Pruning removes negative evidence}: EvolveR discards low-scoring principles, but rare cautionary principles are exactly the ones that would save the agent on the rare task where they apply. \emph{Graph pollution}: InsightGraph creates SIMILAR\_CONCEPT edges aggressively, and a handful of generic insights (\eg, ``use site-restricted search'') become hubs that dominate graph expansion regardless of the query.

\subsection{A Qualitative Lens: Task Surface and Transfer Requirements}
\label{appendix:theoretical_lens}

The failure modes above arise because each provider exposes a different unit of transfer. Similarity-retrieval providers favor memories whose task descriptions are close to the current query; structural-abstraction providers expose reusable procedures with task-specific details removed; and insight-accumulation providers retrieve smaller fragments that can be combined. None of these contracts is uniformly preferable: their utility depends on what the current task needs and on the content that the populated store can actually return.

This perspective motivates content-aware probing without assuming that the router explicitly recovers a latent logical form. Probe snippets reveal the concrete memory available from each provider for the current task, allowing the router to compare provider outputs rather than relying only on static provider descriptions or task-level surface similarity.

\subsection{Case Studies: Complementary Transfer Patterns in Storage}
\label{sec:bridges}

We illustrate these complementary transfer patterns with five cases drawn from the populated memory stores. The cases show how a provider-specific representation can expose useful content that another provider misses; Case~5 presents the reverse situation, in which surface proximity misleads a similarity retriever.

\paragraph{Case 1: Authoritative-PDF retrieval (InsightGraph bridges unrelated surfaces).}
\begin{itemize}[leftmargin=*]
    \item \textbf{Task A}: \emph{``In Emily Midkiff's June 2014 article in a journal named for one of Hreidmar's sons\ldots, what word was quoted from two different authors in distaste for the nature of dragon depictions?''}
    \item \textbf{Task B}: \emph{``What was the volume in m${}^3$ of the fish bag calculated in the University of Leicester paper `Can Hiccup Supply Enough Fish to Maintain\ldots'?''}
\end{itemize}
Because the tasks share essentially no surface vocabulary, dense task-to-task retrieval yields no useful cross-task signals. Yet, InsightGraph connects them via a SIMILAR\_CONCEPT edge between
\emph{``Use targeted site- and filetype-restricted searches combining author, date, and journal-name variants to locate authoritative PDFs''} (from Task A) and
\emph{``Use exact-title and site-restricted search queries (quotes and \texttt{site:}) to rapidly locate authoritative PDFs''} (from Task B).
Although the two tasks have different surface vocabulary, both can reuse the same operational pattern: \emph{identify a unique title $\to$ restrict search by site/filetype $\to$ extract one field from the PDF}. Task-level similarity retrieval does not expose this connection because its unit of comparison is the full task description. InsightGraph can expose it because its indexable unit is an atomic insight, and graph expansion propagates that insight from one task into the retrieval context of the other.

\paragraph{Case 2: Workflow reuse across heterogeneous domains (AWM succeeds; Voyager fails).}
A stored AWM workflow from a ``Mercedes Sosa studio albums'' training task abstracts the solution into a generalized skeleton: \emph{define variables $\{$artist\_name, wiki\_url, start\_year, end\_year$\}$ $\to$ retrieve page $\to$ parse structured list $\to$ filter by year range}. During evaluation, two GAIA tasks from entirely different domains (a Wikipedia-page-count question and a five-member World-of-Warcraft group composition question) both benefit from this workflow because they can reuse the same operation pattern: \emph{open target page $\to$ parse structured list $\to$ filter by predicate $\to$ count}. In contrast, Voyager stores the literal narrative of the Mercedes Sosa trajectory (``I searched and crawled\ldots planned a MediaWiki API query\ldots''), which remains tied to the music domain; consequently, its top-1 retrieval for the WoW task yields an unrelated trajectory. This performance gap emerges precisely when the new task requires the \emph{shape} of a workflow rather than the narration of a specific execution.

\paragraph{Case 3: Anti-pattern warning (ExpeL succeeds; AWM is silent).}
A GAIA task asks: \emph{``How many edits were made to the Wikipedia page on Antidisestablishmentarianism\ldots?''}. The natural agent behavior is to copy the revision counter from the page's History UI, which overcounts because of pagination and soft-reverts. This failure mode appears in the training trajectories. ExpeL, which ingests both successes and failures, stored a warning:
\begin{quote}\small
\textit{``When a Wikipedia task asks for the number of `edits' or `revisions', verify whether the counter in the page history UI is paginated; do not rely directly on the history-page counter.''}
\end{quote}
AWM discards this kind of cautionary note by design (success-only ingestion), so its retrieval on this task cannot prevent the pitfall. The task is solved by ExpeL, InsightGraph, and Agent-KB, but not by AWM, Voyager, DILU, or the no-memory baseline.

\paragraph{Case 4: Executable-skill generalization (SkillWeaver's all-or-nothing payoff).}
SkillWeaver stored a Python function \texttt{find\_adjacent\_players(team, as\_of\_date)} that returns players with jersey numbers immediately before and after a target player on a given date. An evaluation task asking for ``pitchers with the number before and after Taish\={o} Tamai's number as of July 2023'' matches this function signature \emph{exactly}; the skill runs deterministically and returns the answer. A sibling Excel-map navigation task that asks for adjacent cells has no matching skill, and SkillWeaver provides no useful memory. The case reveals the trade-off of executable-skill memory: when the abstract algorithm matches, the skill is strictly better than any workflow description because it \emph{executes}; when it does not, the skill provides no benefit.

\paragraph{Case 5: Similarity trap (DILU misleads on near-paraphrase).}
The converse failure arises when two tasks share surface form but differ in the constraint they impose:
\begin{itemize}[leftmargin=*]
    \item \textbf{Stored trajectory}: \emph{``How many studio albums did Mercedes Sosa release \textbf{between 2000 and 2009}?''}
    \item \textbf{Eval task}: \emph{``How many studio albums did Mercedes Sosa release \textbf{before 2000}?''}
\end{itemize}
The two queries are near-paraphrases in cosine space. DILU returns the stored trajectory with high confidence, and the agent copies its year-range filter \texttt{[2000, 2009]} into the new query, yielding an off-by-decade answer. AWM, which exposes \texttt{\{start\_year, end\_year\}} as placeholders, avoids the trap because placeholders must be re-bound from the current query before execution. This case is the mirror of Cases 1--2: when surface is \emph{too} close to structure, similarity retrieval overfits; structural abstraction always re-binds variables.

\subsection{Implications}
\label{appendix:implications}

These cases illustrate three empirical properties of the provider pool. \emph{(i)} A provider's representation and retrieval contract determine which parts of past experience are exposed for reuse. \emph{(ii)} Surface similarity can be useful, but it can also miss transferable procedures or retrieve a near-paraphrase with incompatible constraints; structural abstractions and smaller insight units offer complementary alternatives. \emph{(iii)} Content-aware probing lets MemAgent inspect what each populated provider can return for the current task and select among these complementary outputs. The analysis explains provider heterogeneity without assuming explicit latent-structure recovery.